\documentclass[10pt,twocolumn,letterpaper]{article}

\usepackage{cvpr}              %

\usepackage{algorithm}
\usepackage{algpseudocode}
\usepackage{colortbl}
\definecolor{oursrow}{RGB}{226,239,250}

\usepackage{lineno}
\definecolor{cvprblue}{rgb}{0.21,0.49,0.74}
\usepackage[pagebackref,breaklinks,colorlinks,allcolors=cvprblue]{hyperref}

\def\confName{CVPR}
\def\confYear{2027}

\title{WAM-Policy: World Action Models with On-Policy Distillation for\\Quadrupedal Loco-Manipulation}

\author{
Junlin Long$^{1*}$ \quad
Xu Deng$^{2*}$ \quad
Yiran Wang$^{1}$ \quad
Yue Yang$^{3\dag}$ \quad
Yang Zhao$^{4\dag}$ \quad
Zeyu Zhang$^{3\ddag}$ \\
$^{1}$Australian Centre for Robotics, USYD \quad
$^{2}$UNSW Sydney \quad
$^{3}$Maincode \quad
$^{4}$La Trobe University \\
\small $^*$Equal contribution.
$^\dag$Corresponding authors: yue@maincode.com, yangzhao.au@outlook.com.
$^\ddag$Project lead.
}

\begin{document}
\twocolumn[{%
\renewcommand\twocolumn[1][]{#1}%
\maketitle
\begin{center}
\vspace{-2em}
  \captionsetup{type=figure,skip=4pt}%
  \centering
  \includegraphics[width=\textwidth]{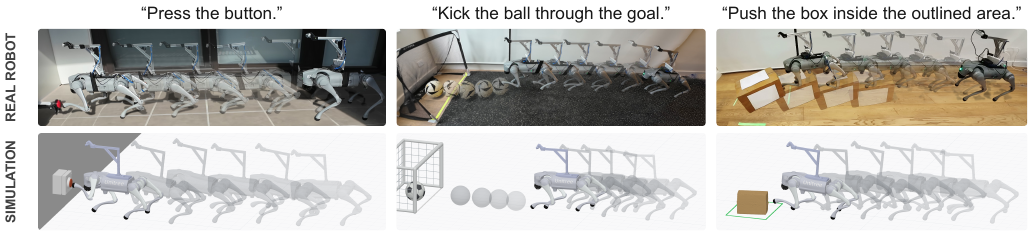}
  \caption{\textbf{WAM-Policy} is a language-conditioned policy for arm-free quadrupedal loco-manipulation, in which a front leg serves as the end-effector.
  Top: the OPSD student (Stage~1) on three tasks on the real robot, as time-lapse composites of one rollout each with intermediate frames translucent.
  Bottom: closed-loop rollouts of the same student on the same tasks in simulation, rendered from the recorded states with earlier poses translucent and spaced out to the right; only the final pose and the objects are at their recorded positions.}
  \label{fig:teaser}
\end{center}%
\vspace{-6pt}%
}]

\patchcmd{\endabstract}{\vspace*{12pt}}{\par\vfill\newpage}{}{\PackageWarning{main}{could not patch the end of the abstract}}
\begin{abstract}
Arm-free quadrupeds can perform loco-manipulation, walking on three legs while manipulating with a front leg, yet no existing policy for it follows language instructions.
World action models (WAMs) have emerged as effective language-conditioned manipulation policies, but are pretrained for robot arms and sample slowly.
WAM-Policy turns one WAM into a quadrupedal loco-manipulation policy via two stages.
The first, Cross-Morphology Policy Adaptation and Acceleration, adapts the arm-centric WAM to quadrupeds and distills the transferred policy into a student with one video and two action steps by on-policy self-distillation (OPSD).
This student, however, is bounded by its teacher, which only imitates demonstrations.
The second stage, Task-Specific Improvement and Merging, therefore tunes the teacher by reinforcement learning (RL) and merges the slow specialists into one few-step student by multi-teacher on-policy distillation (M-OPD).
On QLM-Bench, ten simulated tasks with 1,000 demonstrations synthesized without teleoperation, the transferred policy reaches 77.0\%, and the OPSD student keeps 76.7\% while sampling 14 times faster.
Task-specific RL and M-OPD lift it to 85.0\%.
We also deploy the OPSD student on a real Unitree Go2 Air.
WAM-Policy thus achieves a stronger quadrupedal policy at a fraction of the inference cost.
\href{https://github.com/AIGeeksGroup/WAM-Policy}{Code} \& 
\href{https://aigeeksgroup.github.io/WAM-Policy}{Website}.
\end{abstract}

\section{Introduction}
\label{sec:intro}

Quadrupeds reach terrain that wheeled and fixed-base robots cannot, and language is a natural way to instruct them.
To manipulate, most quadrupeds carry an arm or a gripper~\cite{liu2024vbc,ha2024umilegs,wu2024doggybot}, but an arm-free quadruped can also walk on three legs and use a front leg as the end-effector~\cite{cheng2023legsmanip,arm2024pedipulate,cheng2025rambo}.
Manipulating with a leg spares the robot the weight and mechanical complexity of an added arm~\cite{arm2024pedipulate}.
Existing visual planners for such leg manipulation learn from point clouds and cannot map language instructions to actions end to end~\cite{he2024visualmaniplegs,he2024legmanip}, and, to our knowledge, no demonstrations with language instructions exist for this setting.
Meanwhile, world action models (WAMs), which jointly predict future observations and actions, have emerged as effective language-conditioned manipulation policies~\cite{wu2023gr1,cheang2024gr2,guo2024pad,zhu2025uwm,kim2026cosmospolicy,ye2026dreamzero,li2026lingbotva}.
This paper brings a WAM to arm-free quadrupedal loco-manipulation (\cref{fig:teaser}).

This raises two challenges.
\emph{First, pretrained WAMs are arm-centric, and their multi-step sampling is slow.}
Our base WAM, LingBot-VA~\cite{li2026lingbotva}, sees head and wrist cameras and emits a 30-dimensional dual-arm command, whereas the cameras of a quadruped move with its body and its manipulating leg also supports it.
Its video and action streams are each sampled over many steps at one network function evaluation (NFE) per step, so our transferred policy needs 2.226\,s for a chunk that covers 0.32\,s of robot time, and sampling with one video and two action steps lowers its success.
On the real robot, the body drifts while the whole-body controller~\cite{cheng2025rambo} balances it during sampling, so actions computed from the state before sampling no longer fit the state after it.
\emph{Second, the distilled policy remains bounded by imitation.}
A distilled student is bounded by its teacher, which only imitates demonstrations and still fails on some hard tasks.
Reinforcement learning with task-specific rewards can go beyond the demonstrations, but it tunes the slow high-NFE policy one task at a time, so it yields several slow specialists that must be merged into one policy and accelerated again.

WAM-Policy turns one WAM into a quadrupedal loco-manipulation policy via two stages, one for each challenge: Cross-Morphology Policy Adaptation and Acceleration (Stage~1) and Task-Specific Improvement and Merging (Stage~2).
Both stages end by distilling a high-NFE teacher, the policy itself in Stage~1 and the task-routed teachers in Stage~2, on-policy into a 1/2-NFE student (one video and two action steps), so the final policy is stronger than the transferred one at a fraction of its inference cost.
Our contributions are:
\begin{itemize}
\item \textbf{Cross-morphology adaptation and acceleration.} Stage~1 replaces only the visual and action interfaces of a WAM, fine-tunes it on quadruped demonstrations, and accelerates it by a few-step form of on-policy self-distillation (OPSD)~\cite{zhao2026opsd}, in which the same network with more sampling steps labels the student's own samples.
The transferred policy reaches 77.0\%, and the OPSD student keeps 76.7\% (69.3\% without distillation) while sampling 14 times faster.
\item \textbf{Task-specific improvement and merging.} Stage~2 tunes specialist teachers by per-task reinforcement learning and merges them by multi-teacher on-policy distillation (M-OPD)~\cite{ni2026multiteacher}, which routes each context to the teacher of its task.
On pushing a box and opening a door, the specialists succeed in every episode, against 47\% and 53\% for the OPSD teacher, and the M-OPD student reaches 90\% and 87\%, raising the ten-task success from 76.7\% to 85.0\%, above the OPSD teacher (77.0\%).
\item \textbf{QLM-Bench}, ten simulated loco-manipulation tasks with 1,000 language-annotated demonstrations synthesized without teleoperation, and evaluation in simulation and on a real Unitree Go2 Air, where the OPSD student runs without real-robot data.
\end{itemize}

\section{Related Work}
\label{sec:related}

\paragraph{Quadrupedal loco-manipulation.}
Quadrupeds that follow language mostly pair a VLA/VLM with a controller for velocity, gait, or navigation~\cite{ding2024quarvla,tong2025quartonline,zhao2025more,cheng2025navila}, while manipulation systems attach an arm or gripper~\cite{liu2024vbc,ha2024umilegs,wu2024doggybot,minelli2026tabletop}, sequence skills via an LLM~\cite{ouyang2024longhorizonllm}, or track foot targets with leg controllers~\cite{cheng2023legsmanip,arm2024pedipulate,cheng2025rambo}.
Visual leg-manipulation planners learn from point clouds by reinforcement learning~\cite{he2024visualmaniplegs} or imitation (Leg-Manip~\cite{he2024legmanip}) without language; WAM-Policy is a single language-conditioned RGB policy.

\paragraph{World action models.}
World action models (WAMs) jointly predict future scenes and actions using video-pretrained priors~\cite{wu2023gr1,cheang2024gr2,guo2024pad,zhu2025uwm,kim2026cosmospolicy,ye2026dreamzero}, including our base model LingBot-VA~\cite{li2026lingbotva}.
Prior WAMs target fixed-base arms, humanoids~\cite{zheng2026motionwam,li2026omega0,li2026wholebodywamwbc,zhang2026wholebodywam,qin2026wbwam}, or an armed wheeled-legged quadruped~\cite{ma2026decowam}, and quadruped world models address locomotion or navigation~\cite{wu2022daydreamer,lai2024wmp,bar2025nwm}; none manipulates with a load-bearing leg.
Rather than training a cross-embodiment policy~\cite{oxe2024openx,doshi2024crossformer,wang2024hpt}, we adapt one arm-centric WAM---to our knowledge the first WAM for arm-free quadrupedal loco-manipulation.

\paragraph{Few-step WAMs and on-policy distillation.}
Multi-step joint denoising makes WAMs strong but slow~\cite{li2026lingbotva,kim2026cosmospolicy,ye2026dreamzero}; unlike methods that skip full future prediction~\cite{yuan2026fastwam,zheng2026motionwam,ma2026dit4dit,li2025uva,lv2026realtimewam,su2026onestepwam,zhang2026oswam}, we accelerate the latent plan so that the teacher can supervise it.
Few-step distillation~\cite{salimans2022progressive,song2023consistency,luo2023lcm,yin2024dmd,yin2024dmd2,liu2023rectifiedflow,prasad2024consistencypolicy,wang2025onedp} has been applied offline to WAMs by Flash-WAM~\cite{akbari2026flashwam} and concurrent DirectWAM~\cite{zheng2026directwam} and AnyStep-WAM~\cite{wang2026anystepwam}, whereas on-policy training supervises a student on its own samples~\cite{ross2011dagger,bengio2015scheduled,agarwal2024gkd,huang2025selfforcing}.
Existing on-policy WAM/VLA distillation either labels a single rollout batch~\cite{yang2026wamopd}, preserves the step count~\cite{liu2026wamopd,ni2026multiteacher}, or runs on-policy only along the denoising chain~\cite{oh2026fastopd}.
Whereas prior on-policy self-distillation grants the teacher privileged context~\cite{zhao2026opsd,liu2026opsdv}, our OPSD teacher uses the same closed-loop context with more sampling steps, distilling the student to one video and two action steps along its own denoising chain.

\paragraph{Reinforcement learning and policy merging.}
Reinforcement learning fine-tunes diffusion and flow models~\cite{liu2025flowgrpo,zheng2026diffusionnft} and policies~\cite{ren2024dppo,zhang2025reinflow}, including VLAs~\cite{chen2025pirl,li2025simplevlarl}, and lets a policy go beyond its demonstrations; it is best used to adjust an already competent policy~\cite{huang2026rlposition}.
Tuning one task at a time, however, yields several specialists; Multi-Teacher OPD~\cite{ni2026multiteacher} distills several teachers into one student but keeps their step count, whereas \mbox{M-OPD} merges the specialists into one few-step student.

\section{Method}
\label{sec:method}

\begin{figure*}[t]
  \centering
  \includegraphics[width=\textwidth]{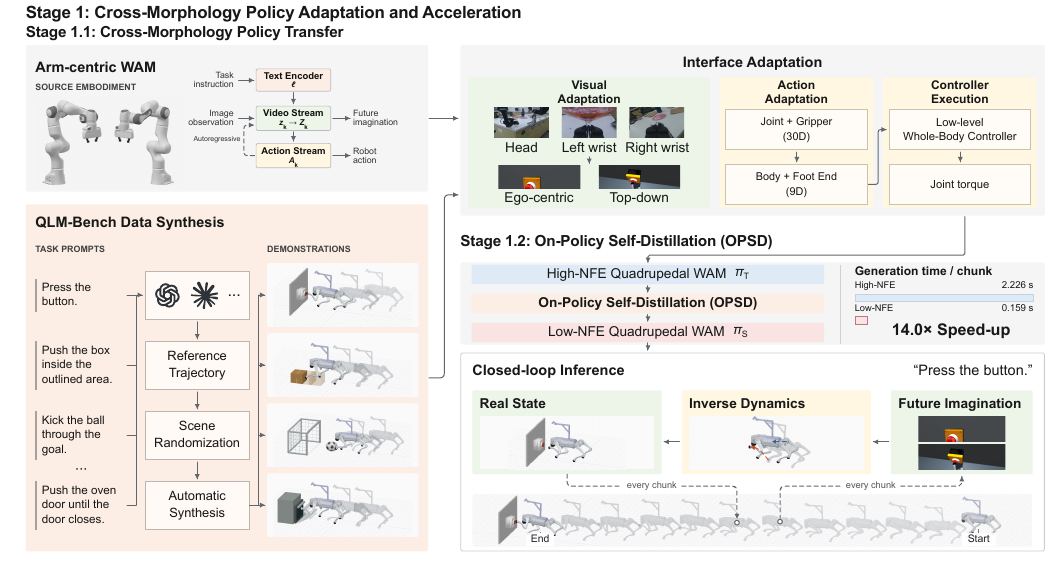}
  \caption{\textbf{Stage~1: Cross-Morphology Policy Adaptation and Acceleration.}
  Stage~1.1 (\cref{sec:transfer}) fine-tunes an arm-centric WAM on quadruped demonstrations synthesized in QLM-Bench (\cref{sec:qlm}).
  Visual Adaptation replaces the head and wrist views (images from Mobile ALOHA~\cite{fu2024mobilealoha}) by the ego-centric and top-down views, Action Adaptation replaces the 30-dimensional arm command by the 9-dimensional quadruped command, and the whole-body controller $\mathcal{C}$ converts each command into joint torques.
  Stage~1.2, OPSD (\cref{sec:opd}), distills the transferred policy, as the high-NFE teacher $\pi_{\mathrm{T}}$, into the 1/2-NFE student $\pi_{\mathrm{S}}$; the generation time counts the sampling of video and action only.
  Bottom right shows the closed loop of the policy on the task ``Press the button.''
  Imagined views and rendered demonstrations are illustrations.}
  \label{fig:transfer}
\end{figure*}

\begin{figure*}[t]
  \centering
  \includegraphics[width=\textwidth]{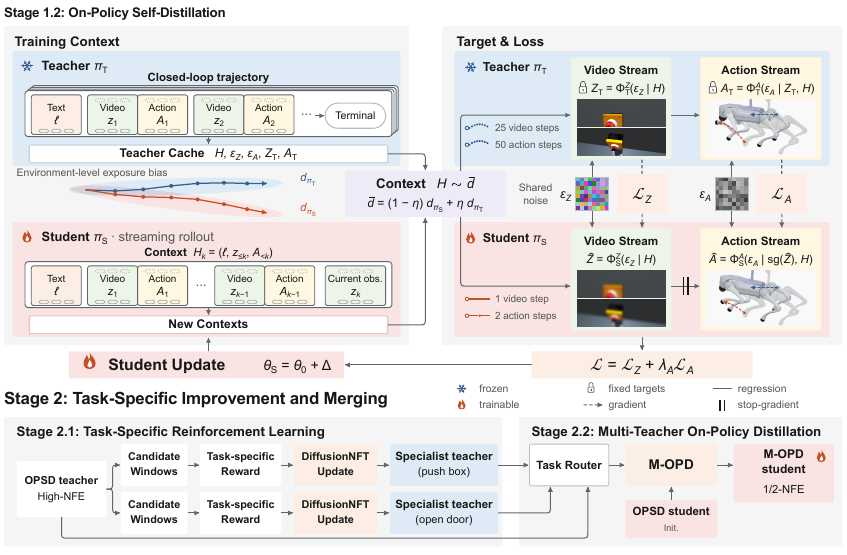}
  \caption{\textbf{On-Policy Self-Distillation (Stage~1.2, top) and Stage~2 (bottom).}
  Top left (Training Context): training contexts mix the teacher cache, which the frozen teacher collects in closed loop before training, with the contexts that the student reaches in its own streaming rollouts ($H\sim\bar d$).
  Top right (Target \& Loss): on each context, teacher and student sample from the same noise, and the student's one-step video stream and two-step action stream are regressed to the targets of the teacher's high-NFE sampler, with a stop-gradient on the plan that conditions the student's action.
  Bottom: Stage~2.1 tunes the OPSD teacher by reinforcement learning into a specialist teacher for each selected task (\cref{sec:rl}), and Stage~2.2, M-OPD, distills the teachers, routed by task, into one 1/2-NFE student that starts from the OPSD student (\cref{sec:mopd}).
  Plan images and rendered actions are illustrations, not model outputs.}
  \label{fig:opd}
\end{figure*}

WAM-Policy turns a pretrained arm-centric WAM into a quadrupedal loco-manipulation policy $\pi_\theta$ in two stages (\cref{sec:stage1,sec:stage2}, \cref{fig:transfer,fig:opd}); \cref{sec:qlm} presents QLM-Bench, which provides the demonstrations and the evaluation tasks.

\paragraph{Notation.}
At chunk $k$, a WAM conditions on the context $H_k=(\ell,\,z_{\le k},\,A_{<k})$: the instruction $\ell$, the latents $z$ of the RGB observations under a frozen causal VAE, and the executed action chunks.
In our base model LingBot-VA~\cite{li2026lingbotva}, a transformer with parameters $\theta$, the video stream predicts the next latent, the plan $Z_k$, and the action stream infers the chunk $A_k$ from it by inverse dynamics.
Both streams are trained with flow matching and sampled from noise $\epsilon_Z,\epsilon_A\sim\mathcal{N}(0,I)$ by Euler integration at one NFE per step (\cref{sec:prelim}).
With samplers $\Phi^Z_\theta$ and $\Phi^A_\theta$ of $N_v$ and $N_a$ steps, $Z_k=\Phi^Z_\theta(\epsilon_Z\mid H_k)$ and $A_k=\Phi^A_\theta(\epsilon_A\mid Z_k,H_k)$ define the WAM policy $\pi_\theta(A_k\mid H_k)$, and executing $A_k$ yields $z_{k+1}$ and $H_{k+1}$.

\subsection{Stage 1: Cross-Morphology Policy Adaptation and Acceleration}
\label{sec:stage1}

\subsubsection{Cross-Morphology Policy Transfer}
\label{sec:transfer}

LingBot-VA is pretrained for arms: its video stream sees the head and wrist cameras of a manipulator, and its action stream emits a 30-dimensional dual-arm command~\cite{li2026lingbotva}.
We replace only these two interfaces and keep the rest of the model, including the frozen VAE and text encoder (Stage~1.1 in \cref{fig:transfer}).
We call this transfer cross-morphology rather than cross-embodiment, which in manipulation often means a different arm or gripper: our target is an arm-free quadruped that manipulates with one of its legs.
Fine-tuning both streams with the new interfaces on the QLM-Bench demonstrations (\cref{sec:qlm}) with flow matching gives the transferred policy $\pi_{\theta_0}$, one policy for all ten tasks, still with the high-NFE sampler.

\paragraph{Visual adaptation.}
The robot carries two body-mounted RGB cameras: a wide-angle ego camera at the head looks forward, and a top-down camera, raised above the trunk and pitched down, looks at the workspace.
The two views are encoded by the frozen VAE and concatenated spatially, as LingBot-VA does for its views, so $z_k$ and $Z_k$ cover both.
Because both cameras move with the body, predicting the next latent requires modeling how the robot's own motion changes both views.

\begin{figure*}[t]
  \centering
  \includegraphics[width=\textwidth]{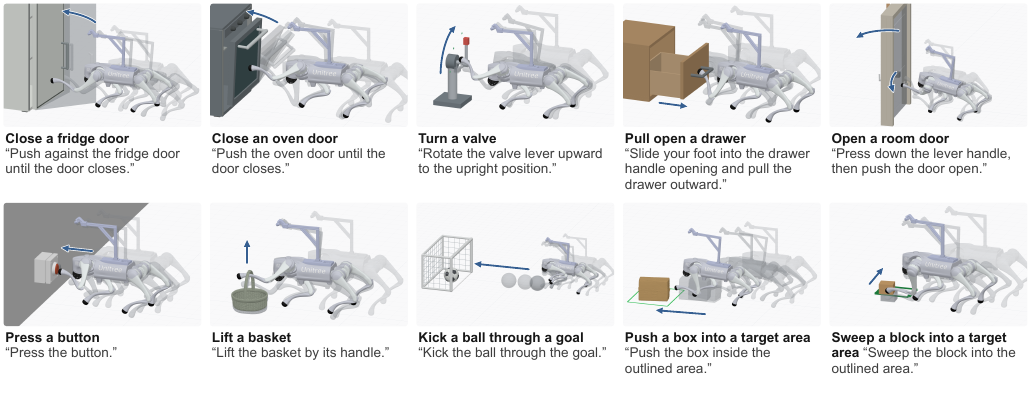}
  \caption{\textbf{The ten tasks of QLM-Bench.}
  One successful rollout per task, with earlier poses translucent; arrows mark the intended motion.}
  \label{fig:qlm_tasks}
\end{figure*}

\paragraph{Action adaptation.}
We replace the arm action space with the command interface of RAMBO~\cite{cheng2025rambo}, a whole-body controller $\mathcal{C}$ for a quadruped that walks on three legs and manipulates with its front-left leg.
A command $a\in\mathbb{R}^9$ specifies the planar base velocity and yaw rate, the target position of the manipulating foot, and its target contact force.
The action stream outputs a chunk $A_k$ of such commands, and only its input embedding and output projection are re-initialized.
$\mathcal{C}$ stays fixed and converts each command into joint torques; the term policy refers only to $\pi_\theta$.

\subsubsection{On-Policy Self-Distillation}
\label{sec:opd}

The high-NFE sampler makes $\pi_{\theta_0}$ too slow for real time.
We take it as the teacher $\pi_{\mathrm{T}}$, with $\theta_0$ frozen and its original sampler, and distill it in a single training phase into a student $\pi_{\mathrm{S}}$ with $\theta_{\mathrm{S}}=\theta_0+\Delta$, where the LoRA update~\cite{hu2022lora} $\Delta$ is the only trained parameter, and with its sampler reduced directly to $N_v/N_a=1/2$ (\cref{fig:opd}, top); $\Phi^Z_{\mathrm{T}},\Phi^A_{\mathrm{T}}$ and $\Phi^Z_{\mathrm{S}},\Phi^A_{\mathrm{S}}$ denote their samplers.
As in On-Policy Self-Distillation (OPSD)~\cite{zhao2026opsd}, the policy is distilled from itself, but our teacher sees the same context as the student and is ahead through its high-NFE sampler rather than through additional context.

\paragraph{Targets and loss.}
For each training context $H$, we draw $(\epsilon_Z,\epsilon_A)$ once, keep it with the context, and compare the two deterministic samplers on the same noise,
\begin{equation}
\begin{aligned}
\hat Z&=\Phi^Z_{\mathrm{S}}(\epsilon_Z\mid H), & \hat A&=\Phi^A_{\mathrm{S}}(\epsilon_A\mid \mathrm{sg}(\hat Z),H),\\
Z_{\mathrm{T}}&=\Phi^Z_{\mathrm{T}}(\epsilon_Z\mid H), & A_{\mathrm{T}}&=\Phi^A_{\mathrm{T}}(\epsilon_A\mid Z_{\mathrm{T}},H).
\end{aligned}
\label{eq:pairs}
\end{equation}
The target chunk $A_{\mathrm{T}}$ is conditioned on the teacher's own plan and is thus the chunk the teacher would execute, whereas the student's chunk $\hat A$ is conditioned on the student's own plan through a stop-gradient $\mathrm{sg}(\cdot)$.
Being independent of $\theta_{\mathrm{S}}$, the targets can be cached, and the student is trained with
\begin{equation}
\begin{aligned}
\mathcal{L}&=\mathcal{L}_Z+\lambda_A\mathcal{L}_A,\\
\mathcal{L}_Z&=\bar\rho_c(\hat Z-Z_{\mathrm{T}}),\quad \mathcal{L}_A=\bar\rho_c(\hat A-A_{\mathrm{T}}),
\end{aligned}
\label{eq:loss}
\end{equation}
where $\bar\rho_c$ averages the pseudo-Huber distance $\rho_c(r)=\sqrt{r^2+c^2}-c$~\cite{song2024ict} over the valid entries, and $\lambda_A$ weights the action term.

\paragraph{Training contexts.}
\label{sec:exposure}
Replacing the teacher with the student brings two mismatches (\cref{sec:prelim_distill}).
At the generation level, the student samples through its own short chain and conditions its action on its own plan; at the environment level, the resulting errors change its actions, so it visits contexts from $d_{\pi_{\mathrm{S}}}$ rather than $d_{\pi_{\mathrm{T}}}$, where $d_\pi$ is the distribution of contexts that $\pi$ visits in closed loop~\cite{ross2011dagger,bengio2015scheduled}.
\Cref{eq:pairs} handles the first on any context; for the second, training contexts are drawn from a mixture of the two closed-loop distributions,
\begin{equation}
\min_{\Delta}\;\mathbb{E}_{H\sim\bar d}\,\mathbb{E}_{\epsilon_Z,\epsilon_A}\big[\mathcal{L}\big],\qquad
\bar d=(1-\eta)\,d_{\pi_{\mathrm{S}}}+\eta\,d_{\pi_{\mathrm{T}}}.
\label{eq:objective}
\end{equation}
Streaming contexts sample $d_{\pi_{\mathrm{S}}}$ from parallel closed-loop rollouts of $\pi_{\mathrm{S}}$ with its latest weights, labeled by the teacher at the student's chunk boundaries; the teacher cache samples $d_{\pi_{\mathrm{T}}}$ from rollouts of $\pi_{\mathrm{T}}$ before training.
\Cref{tab:abl_contexts} varies $\eta$, and \cref{alg:opd} in the supplementary material summarizes the procedure.

\subsection{Stage 2: Task-Specific Improvement and \texorpdfstring{\mbox{Merging}}{Merging}}
\label{sec:stage2}

\subsubsection{Task-Specific Reinforcement Learning}
\label{sec:rl}

The OPSD student is bounded by its teacher, which only imitates the demonstrations, so Stage~2.1 tunes the transferred policy itself by reinforcement learning, one task at a time, adjusting a policy that imitation has already made competent rather than learning one from scratch~\cite{huang2026rlposition}.
For a task $g$, only a LoRA update $\Delta_g$ on top of $\theta_0$ is trained, which gives a specialist teacher $\pi^{g}_{\mathrm{T}}$ with the original high-NFE sampler.
We tune two tasks, pushing the box and opening the door, with a window-level variant of DiffusionNFT~\cite{zheng2026diffusionnft}.

From states of the teacher's own closed-loop rollouts, a slowly updated copy of the policy runs groups of candidate windows of several chunks, each with one noise pair kept over the window.
Each window receives one reward computed only from the trajectory of the task object or of the foot relative to it: how far the box advances without deviating from a straight push, and how far toward the door the foot rests on the handle after contact.
The DiffusionNFT loss pulls the policy toward the candidates rewarded above the mean of their start and away from those below it (\cref{sec:suppl_rl}).

\begin{table*}[t]
  \centering
  \footnotesize
  \caption{\textbf{Success rate (\%) on QLM-Bench Standard100.}
  WAM-Policy rows use 300 episodes (100 cases $\times$ 3 policy seeds).
  $\pm$ is the standard deviation of the ten-task mean over policy seeds 42, 43, and 44, and Seeds gives the successes out of 100 episodes for each.
  NFE gives the video/action sampling steps.
  Time is the per-chunk sampling time of the two streams on an RTX 5070 Ti, without encoding the observations and with the LoRA update of each student merged (\cref{sec:suppl_compute}); a chunk covers 0.32\,s of robot time.
  Leg-Manip is averaged over three training seeds, each evaluated on the 100 cases with evaluation seed 42, so it has no results per seed.
  Each specialist teacher is evaluated on all ten tasks but used by M-OPD only on its own task (\cref{eq:mopd}, \cref{sec:suppl_impl}).
  Shaded rows are our method.}
  \label{tab:qlm}
  \setlength{\tabcolsep}{2.5pt}
  \resizebox{\linewidth}{!}{%
    \begin{tabular}{lcc*{10}{c}cc}
      \toprule
      Method & NFE & Time (s) & Fridge & Oven & Valve & Drawer & Door & Press & Basket & Kick & Push & Sweep & All & Seeds \\
      \midrule
      Leg-Manip~\cite{he2024legmanip} & -- & -- & 93 & 90 & 93 & 70 & 0 & 80 & 80 & 87 & 17 & 7 & 61.7 & -- \\
      \midrule
      \multicolumn{15}{l}{\emph{Stage~1: OPSD}} \\
      Teacher & 25/50 & 2.226 & 100 & 100 & 100 & 97 & 53 & 77 & 100 & 97 & 47 & 0 & 77.0$\pm$4.0 & 77/73/81 \\
      Student w/o distillation & 1/2 & 0.156 & 100 & 100 & 47 & 0 & 93 & 93 & 100 & 97 & 60 & 3 & 69.3$\pm$1.2 & 68/70/70 \\
      \rowcolor{oursrow}
      \textbf{Student (Ours)} & 1/2 & 0.159 & 100 & 100 & 100 & 97 & 37 & 83 & 100 & 100 & 50 & 0 & 76.7$\pm$4.7 & 82/75/73 \\
      \midrule
      \multicolumn{15}{l}{\emph{Stage~2: M-OPD}} \\
      Specialist teacher (push box) & 25/50 & 2.746 & 100 & 100 & 87 & 70 & 63 & 90 & 100 & 93 & 100 & 10 & 81.3$\pm$2.5 & 84/79/81 \\
      Specialist teacher (open door) & 25/50 & 2.740 & 100 & 100 & 97 & 87 & 100 & 83 & 100 & 100 & 43 & 53 & 86.3$\pm$6.5 & 80/93/86 \\
      \rowcolor{oursrow}
      \textbf{Student (Ours)} & 1/2 & 0.159 & 100 & 100 & 97 & 93 & 87 & 83 & 100 & 100 & 90 & 0 & 85.0$\pm$2.6 & 84/83/88 \\
      \bottomrule
    \end{tabular}%
  }
\end{table*}

\subsubsection{Multi-Teacher On-Policy Distillation}
\label{sec:mopd}

Stage~2.2 merges the specialists into one 1/2-NFE student for all tasks by on-policy distillation from several teachers~\cite{ni2026multiteacher}, which we route by task.
Let $\mathcal{G}$ be the set of tasks tuned in Stage~2.1 and $g(\ell)$ the task of the instruction $\ell$ in a context $H$.
The routed teacher is
\begin{equation}
\pi_{\mathrm{M}}(\cdot\mid H)=
\begin{cases}
\pi^{g(\ell)}_{\mathrm{T}}(\cdot\mid H), & g(\ell)\in\mathcal{G},\\
\pi_{\mathrm{T}}(\cdot\mid H), & \text{otherwise},
\end{cases}
\label{eq:mopd}
\end{equation}
and M-OPD optimizes \cref{eq:objective} with $\pi_{\mathrm{T}}$ replaced by $\pi_{\mathrm{M}}$: the routed teacher gives the targets of \cref{eq:pairs} on every context, and each teacher fills the teacher cache only with closed-loop contexts of its own tasks.
The student continues from the OPSD student, so $\Delta$ starts from its update instead of zero, and keeps $N_v/N_a=1/2$.
As in OPSD, half of the contexts of an update come from the student's own rollouts, so the routed teacher also labels the contexts that the student passes through while it moves from its Stage~1 behavior toward a specialist.

\section{QLM-Bench}
\label{sec:qlm}

QLM-Bench (Quadrupedal Loco-Manipulation Benchmark) is a MuJoCo benchmark for language-conditioned loco-manipulation in which a Unitree Go2 manipulates with its front-left leg, with the two cameras, the commands, and the whole-body controller $\mathcal{C}$ of \cref{sec:transfer}.
The benchmark has ten tasks (\cref{fig:qlm_tasks}): close a fridge door, close an oven door, turn a valve, pull open a drawer, open a room door, press a button, lift a basket, kick a ball through a goal, push a box into a target area, and sweep a block into a target area.

\paragraph{Demonstration synthesis.}
For each task, a large multimodal model drafts a reference trajectory of commands from a short description of the task (\cref{fig:transfer}, bottom left).
After human review, the reference is retargeted to randomized scenes and executed by $\mathcal{C}$, and the successful episodes are kept as demonstrations, with neither teleoperation nor hand-designed trajectories.
This gives 1,000 demonstrations, 100 per task, each recording the two camera streams, the executed commands, and a language instruction.
We will release the benchmark and the demonstrations with our code.

\section{Experiments}
\label{sec:exp}

The experiments test that Stage~1 yields a competent quadruped policy (\cref{sec:exp_transfer}) that samples faster than real time without losing this competence (\cref{sec:exp_distill}). They then test that Stage~2.1 yields specialist teachers that outperform the OPSD teacher on their own tasks, that Stage~2.2 merges them into one 1/2-NFE student that keeps most of their gains (\cref{sec:exp_stage2}), and that the OPSD student runs on a real robot (\cref{sec:exp_real}).
\Cref{sec:exp_ablation} ablates OPSD and the inputs of the teacher.
Because OPSD does not depend on the embodiment, we also apply it to an arm WAM and ablate it there (\cref{sec:suppl_arm}).

\begin{table*}[t]
  \footnotesize
  \setlength{\tabcolsep}{3pt}
  \begin{minipage}[t]{0.335\textwidth}
    \centering
    \caption{\textbf{Training contexts.} Success rate (\%) after 300 updates; $\pm$ and Seeds as in \cref{tab:qlm}, and $\eta$ as in \cref{eq:objective}.}
    \label{tab:abl_contexts}
    \begin{tabular}{lcc}
    \toprule
    Variant & All & Seeds \\
    \midrule
    Teacher cache only ($\eta=1$) & 77.0$\pm$3.5 & 79/73/79 \\
    Streaming only ($\eta=0$) & 76.3$\pm$4.0 & 80/72/77 \\
    \rowcolor{oursrow}
    \textbf{OPSD (Ours, $\eta=0.5$)} & 76.7$\pm$4.7 & 82/75/73 \\
    \bottomrule
    \end{tabular}
  \end{minipage}\hfill
  \begin{minipage}[t]{0.335\textwidth}
    \centering
    \caption{\textbf{Longer training.} Successes out of 100 episodes with policy seed 42.}
    \label{tab:abl_longer}
    \vspace{1.5pt}

    \begin{tabular}{lcc}
    \toprule
    Variant & Updates & Successes \\
    \midrule
    Teacher cache only ($\eta=1$) & 300 & 79 \\
    Teacher cache only ($\eta=1$) & 2,000 & 77 \\
    \rowcolor{oursrow}
    \textbf{OPSD (Ours, $\eta=0.5$)} & 300 & 82 \\
    \rowcolor{oursrow}
    \textbf{OPSD (Ours, $\eta=0.5$)} & 900 & 79 \\
    \bottomrule
    \end{tabular}
  \end{minipage}\hfill
  \begin{minipage}[t]{0.29\textwidth}
    \centering
    \caption{\textbf{Distillation objective.} Successes out of 100 episodes with policy seed 42; distilled for 900 updates.}
    \label{tab:abl_objective}
    \begin{tabular}{lc}
    \toprule
    Variant & Successes \\
    \midrule
    Student w/o distillation & 68 \\
    Distribution matching ($\eta=0$) & 59 \\
    \rowcolor{oursrow}
    \textbf{OPSD (Ours, $\eta=0.5$)} & 79 \\
    \bottomrule
    \end{tabular}
  \end{minipage}

  \vspace{6pt}
  \begin{minipage}[t]{0.38\textwidth}
    \centering
    \caption{\textbf{Sampling steps without distillation.} Success rate (\%) of the teacher's parameters; $\pm$ and Seeds as in \cref{tab:qlm}.}
    \label{tab:abl_steps}
    \vspace{3.5pt}

    \begin{tabular}{lccc}
    \toprule
    Variant & NFE & All & Seeds \\
    \midrule
    Teacher & 25/50 & 77.0$\pm$4.0 & 77/73/81 \\
    W/o distillation (2/4-NFE) & 2/4 & 78.0$\pm$1.7 & 79/79/76 \\
    Student w/o distillation & 1/2 & 69.3$\pm$1.2 & 68/70/70 \\
    \bottomrule
    \end{tabular}
  \end{minipage}\hfill
  \begin{minipage}[t]{0.34\textwidth}
    \centering
    \caption{\textbf{Proprioception in Stage~1.1.} Success rate (\%) with policy seed 42, without distillation; All is over the ten tasks, and the other seven differ by at most one episode (\cref{tab:suppl_abl_full}).}
    \label{tab:abl_proprio}
    \vspace{2pt}

    \begin{tabular}{lcccc}
    \toprule
    Variant & Valve & Door & Push & All \\
    \midrule
    RGB + proprioception & 0 & 0 & 100 & 67.0 \\
    RGB only (Teacher) & 100 & 100 & 10 & 77.0 \\
    \bottomrule
    \end{tabular}
  \end{minipage}\hfill
  \begin{minipage}[t]{0.25\textwidth}
    \centering
    \caption{\textbf{Real-robot results on the Unitree Go2 Air.}}
    \label{tab:real}
    \begin{tabular}{lccc}
    \toprule
    Method & Press & Kick & Push \\
    \midrule
    \multicolumn{4}{l}{\emph{Success / trials}} \\
    \rowcolor{oursrow}
    \textbf{OPSD student} & 3/10 & 1/10 & 3/10 \\
    \midrule
    \multicolumn{4}{l}{\emph{Completion time (s)}} \\
    \rowcolor{oursrow}
    \textbf{OPSD student} & 37 & 37 & 34 \\
    \bottomrule
    \end{tabular}
  \end{minipage}
\end{table*}

\subsection{Setup}
\label{sec:exp_setup}

\paragraph{Evaluation protocol.}
We evaluate on QLM-Bench (\cref{sec:qlm}) with its Standard100 protocol: ten held-out scene configurations (cases) per task, a time limit of 24 to 60\,s per task, and policy seeds 42, 43, and 44 on each case, for 300 episodes per policy.
A policy seed only sets the initial noise of the policy, while the case fixes the scene, its reset, and the physics.
Simulation waits for the policy at every chunk, so success rates do not depend on inference speed.
Paired McNemar tests between the policies are in \cref{sec:suppl_stats}.

\paragraph{Policies.}
\emph{Teacher} is the transferred policy $\pi_{\theta_0}$ of Stage~1.1 (\cref{sec:transfer}) with its original high-NFE sampler, which we call the OPSD teacher where it appears next to the specialist teachers.
\emph{Student w/o distillation} samples the same parameters ($\Delta=0$) with $N_v/N_a=1/2$.
\emph{Student (Ours)}, which we call the OPSD student, is the 1/2-NFE policy $\theta_0+\Delta$ that OPSD trains on equal shares of streaming and teacher-cache contexts ($\eta=0.5$ in \cref{eq:objective}).
For Stage~2, the \emph{specialist teachers} add the LoRA update $\Delta_g$ of Stage~2.1 to $\theta_0$ and keep the high-NFE sampler (\cref{sec:rl}), and the M-OPD \emph{Student (Ours)} is the 1/2-NFE policy of Stage~2.2 (\cref{sec:mopd}).
We also compare with \emph{Leg-Manip}, our reproduction of the method of He~\etal~\cite{he2024legmanip}.

\subsection{Stage 1: Cross-Morphology Policy Adaptation and Acceleration}
\label{sec:exp_stage1}

\subsubsection{Cross-Morphology Policy Transfer}
\label{sec:exp_transfer}

The transferred policy, the Teacher of \cref{tab:qlm}, succeeds in 77.0\% of the episodes as one language-conditioned RGB policy for all ten tasks.
Leg-Manip, which trains a point-cloud planner for each task on 4 to 80 demonstrations (\cref{sec:suppl_impl}), reaches 61.7\%; opening the door has no demonstrations, so we record 0.

\subsubsection{On-Policy Self-Distillation}
\label{sec:exp_distill}

\paragraph{Speed.}
The teacher takes 2.226\,s to generate a chunk that covers 0.32\,s of robot time, about seven times slower than real time (\cref{tab:qlm}).
With 1/2 NFE, the same parameters take 0.156\,s, 14 times faster and about half the duration of a chunk; \cref{sec:suppl_compute} gives the measurement.

\paragraph{Cost of reducing steps.}
Without distillation, this speed lowers success from 77.0\% to 69.3\% (\cref{tab:qlm}).
The loss comes from pulling the drawer open and turning the valve, which together cost 15 points of the ten-task mean, and is partly offset by opening the door and pressing the button.

\paragraph{Distillation.}
OPSD recovers the success of the teacher at 1/2 NFE, with 76.7\% against 77.0\% (\cref{tab:qlm}).
On the same cases and seeds, it gains 53 episodes and loses 31 against the student without distillation, and restores the two tasks that fewer steps break: pulling the drawer open rises from 0 to 29 of 30 episodes and turning the valve from 14 to 30.

\begin{figure*}[t]
  \centering
  \includegraphics[width=\textwidth]{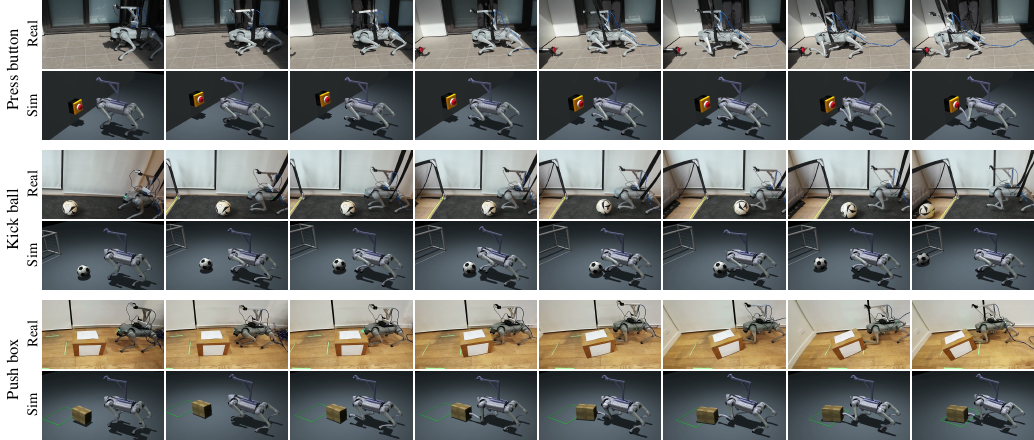}
  \caption{\textbf{Key frames of real-robot and simulated rollouts} of the OPSD student on the three tasks of \cref{tab:real}.
  For each task, the top row (Real) is a rollout on the Unitree Go2 Air and the bottom row (Sim) a rollout in QLM-Bench; time runs from left to right.}
  \label{fig:real_keyframes}
\end{figure*}

\subsection{Stage 2: Task-Specific Improvement and \texorpdfstring{\mbox{Merging}}{Merging}}
\label{sec:exp_stage2}

The OPSD student matches but does not exceed its teacher, so Stage~2.1 tunes the transferred policy on two tasks on which it often fails, pushing the box (47\%) and opening the door (53\%).
Each specialist teacher succeeds in all 30 episodes of its own task (\cref{tab:qlm}); its training curves are in \cref{fig:rl_reward}.
Stage~2.2 merges both by M-OPD into one 1/2-NFE student, which learns the other eight tasks from the OPSD teacher (\cref{sec:mopd}).
The M-OPD student pushes the box in 90\% and opens the door in 87\% of the episodes, against 50\% and 37\% for the OPSD student, so it keeps most of the gain of its specialist teachers; on each of the other eight tasks it differs from the OPSD student by at most one episode, and it still never sweeps the block, which the router does not send to the open-door specialist (53\%; \cref{sec:suppl_impl}).
Through these two tasks alone, its ten-task success rises to 85.0\%, above the OPSD student (76.7\%) and the transferred policy, the OPSD teacher (77.0\%), while it samples a chunk in 0.159\,s instead of 2.226\,s and replaces two specialist teachers at about 2.7\,s each (\cref{tab:qlm}).

\subsection{Ablation}
\label{sec:exp_ablation}

\paragraph{Training contexts.}
We vary the source of the training contexts through $\eta$ (\cref{tab:abl_contexts}): the teacher cache alone already supervises the student on its own samples, and streaming contexts add those that it reaches in closed loop (\cref{sec:opd}).
All three students recover the teacher at 1/2 NFE, with 76.3\% to 77.0\% against 77.0\%, so OPSD is not sensitive to the source of its contexts: the recovery comes from the objective that the three share, the same-noise regression along the student's own sampling chain.
We use the equal mixture by default.

\paragraph{Number of updates.}
We train beyond the default 300 updates to test whether the student keeps improving (\cref{tab:abl_longer}).
With the teacher cache only, 2,000 updates give 77 successes in 100 episodes against 79 after 300, and with the default mixture, 900 updates give 79 against 82.
The distillation loss on held-out contexts keeps falling slowly over these updates (\cref{fig:opsd_loss}), but the closed-loop success does not follow it, so we keep the default 300 updates.

\paragraph{Distillation objective.}
We replace the regression of OPSD by distribution matching~\cite{yin2024dmd,yin2024dmd2}, which trains the student against the difference between the scores of the teacher and of a second model fitted to the student's own outputs, on streaming contexts (\cref{tab:abl_objective}, \cref{sec:suppl_qlm_ablation}).
After 900 updates, this student succeeds in 59 of 100 episodes, below the 68 of the student without distillation, whereas OPSD reaches 79 after the same number of updates.
In this configuration, matching distributions thus does not recover the success that the same-noise regression does.

\paragraph{Sampling steps.}
We reduce the sampling steps of the teacher's parameters without any distillation, to locate where the loss of success begins (\cref{tab:abl_steps}).
With 2/4 NFE the success is 78.0\%, as high as the 77.0\% with 25/50 NFE, and only the last halving to 1/2 NFE lowers it, to 69.3\%.
OPSD recovers this last halving: its student reaches 76.7\% with half the network evaluations of 2/4 NFE, and the same distillation is what merges the specialists in Stage~2.

\paragraph{Proprioception.}
Both cameras are mounted on the body, so their views shake as the robot moves, and we test whether a proprioceptive state reduces the noise that this shaking adds to the video stream.
We add a 10-dimensional proprioceptive state to the inputs and predictions of the teacher, trained otherwise identically (\cref{tab:abl_proprio}, \cref{sec:suppl_qlm_ablation}).
It succeeds in 67 of 100 episodes against 77: it pushes the box in all 10 episodes but neither turns the valve nor opens the door, which the RGB-only teacher does in all 10.
Predicting the state jointly with the video and the action thus lowers success, so WAM-Policy observes RGB only.

\subsection{Real-Robot Deployment}
\label{sec:exp_real}

We deploy the OPSD student on a Unitree Go2 Air with the two cameras of \cref{sec:transfer}: the ego view comes from the Go2's built-in camera and the top-down view from a RealSense D435i.
The command interface and the whole-body controller $\mathcal{C}$ are those of the simulation, and the student runs off-board, with the VAE, on a workstation with a single RTX 5070 Ti that connects to the robot by wired Ethernet.
The policy is trained only on simulated data from QLM-Bench and sees no real-robot data.

We test three tasks with ten trials each, and the OPSD student succeeds in 7 of 30 trials overall (\cref{tab:real}).
\Cref{fig:real_keyframes} compares key frames of real-robot and simulated rollouts of these tasks, and \cref{fig:teaser} (top) shows time-lapse composites of real-robot rollouts.
On this workstation, sampling takes 0.159\,s per chunk, but the closed loop takes 0.66\,s with the encoding of the observations; since this exceeds the 0.32\,s of a chunk, the robot holds its last command between chunks.

\section{Conclusion}
\label{sec:conclusion}

We presented WAM-Policy, which turns an arm-centric world action model into a language-conditioned policy for quadrupedal loco-manipulation.
Stage~1 adapts the WAM to quadrupeds and distills it into a few-step student by OPSD, which is bounded by its teacher.
Stage~2 therefore tunes the transferred policy into specialist teachers by reinforcement learning and merges them by M-OPD into one 1/2-NFE student that is stronger than the transferred policy (85.0\% against 77.0\%) at a fraction of its inference cost.
We further build QLM-Bench, ten quadrupedal loco-manipulation tasks with demonstrations synthesized without teleoperation, and evaluate WAM-Policy in simulation and on a real quadruped.

\section*{LLM Usage Statement}
\label{sec:llm_usage}
We used LLMs for writing, review, figures, code, ideation, and one step of the data pipeline, as listed below.
The authors checked all of the resulting content and take full responsibility for it.
\textbf{Writing.}
The text was drafted and revised with the help of Claude (Anthropic), including wording, editing, and grammar.
Claude also assisted the literature search for \cref{sec:related}, and Claude Code (Anthropic) checked every bibliography entry that has an arXiv identifier against its arXiv record (title and authors).
Before submission, Claude (Anthropic), Codex (OpenAI), and Gemini (Google) reviewed the manuscript for errors and inconsistencies.
\textbf{Figures.}
The drawing code and the layout of \cref{fig:transfer,fig:opd} were generated with the help of Codex (OpenAI), and Claude (Anthropic) edited the SVG source of some of their illustrative plan and view images; as their captions state, these images are illustrations, not outputs of our models.
\Cref{fig:teaser,fig:qlm_tasks} were assembled with the help of Claude Code (Anthropic), and \cref{fig:real_keyframes} with the help of Claude (Anthropic).
The plots and rollout figures of the supplementary material were produced by scripts written with the help of Claude Code (Anthropic), from logged training data and recorded simulation states.
The simulation renders in the figures are made from the recorded states of our own simulation runs.
\textbf{Code and experiments.}
The research code and the scripts that run the experiments and compute the statistics were written with the help of the LLM coding agents Claude Code (Anthropic) and Codex (OpenAI), and the authors reviewed them.
All reported results come from running this code.
\textbf{Data synthesis.}
As described in \cref{sec:qlm}, QLM-Bench uses a large multimodal model, Codex (OpenAI), to draft the reference trajectory of each task, and a human reviews every reference before it is used.
\textbf{Ideation.}
Claude (Anthropic) and Codex (OpenAI) assisted the discussion of the method and the experimental design.

{
    \small
    \bibliographystyle{ieeenat_fullname}
    \bibliography{main}
}

\clearpage
\setcounter{page}{1}
\maketitlesupplementary

\setcounter{section}{0}
\renewcommand{\thesection}{\Alph{section}}
\def\theHsection{suppl.\Alph{section}}
\setcounter{table}{0}
\renewcommand{\thetable}{S\arabic{table}}
\def\theHtable{suppl.\arabic{table}}
\setcounter{figure}{0}
\renewcommand{\thefigure}{S\arabic{figure}}
\def\theHfigure{suppl.\arabic{figure}}
\setcounter{equation}{0}
\renewcommand{\theequation}{S\arabic{equation}}
\def\theHequation{suppl.\arabic{equation}}
\setcounter{algorithm}{0}
\renewcommand{\thealgorithm}{S\arabic{algorithm}}
\def\theHalgorithm{suppl.\arabic{algorithm}}
\setlength{\textfloatsep}{14pt plus 2pt minus 2pt}
\setlength{\dbltextfloatsep}{14pt plus 2pt minus 2pt}

\section{Preliminaries}
\label{sec:prelim}

\subsection{World Action Models}
\label{sec:prelim_wam}

A world action model acts in chunks.
Given a language instruction $\ell$, it conditions on the context $H_k=(\ell,\,z_{\le k},\,A_{<k})$, where $z_k$ is the latent of the RGB observation $x_k$ at the start of chunk $k$ under a frozen causal VAE, and $A_{<k}$ are the action chunks executed so far.
We build on LingBot-VA~\cite{li2026lingbotva}, a transformer with parameters $\theta$ that processes video and action tokens in two streams.
Each stream has its own input embedding and output projection, and the two streams interact through joint self-attention at every layer.
The transformer is initialized from the Wan2.2-5B video generator~\cite{wan2025wan}.
The video stream predicts the latent of the next observation, the latent plan $Z_k$.
The action stream acts as an inverse dynamics model and infers the chunk $A_k$ from this plan.
We write $v^Z_\theta$ and $v^A_\theta$ for their velocity fields.

\paragraph{Training.}
Both streams are trained with flow matching~\cite{lipman2023flowmatching,liu2023rectifiedflow}.
A clean sample $y$ is noised to $y_\sigma=(1-\sigma)\,y+\sigma\epsilon$ with $\epsilon\sim\mathcal{N}(0,I)$, so $\sigma=1$ is pure noise, and the velocity field regresses $\epsilon-y$.
On demonstrations, the video stream learns to predict the latent $z_{k+1}$ of the next observation, and the action stream learns to recover the executed chunk $A_k$ from this real latent (teacher forcing).

\paragraph{Sampling.}
Each stream integrates its flow from $\sigma=1$ to $\sigma=0$ with $N$ Euler steps on a decreasing grid $1=\sigma_0>\sigma_1>\dots>\sigma_N=0$, starting from $y_{\sigma_0}=\epsilon$,
\begin{equation}
y_{\sigma_i} = y_{\sigma_{i-1}}+(\sigma_i-\sigma_{i-1})\,v_\theta(y_{\sigma_{i-1}},\sigma_{i-1}\mid\cdot\,),
\label{eq:euler}
\end{equation}
which costs one network function evaluation (NFE) per step.
We write $\Phi^Z_{\theta,N_v}(\epsilon_Z\mid H)$ and $\Phi^A_{\theta,N_a}(\epsilon_A\mid Z,H)$ for the endpoints of the video and action samplers with $N_v$ and $N_a$ steps, and drop the step counts when they are clear.
For each chunk, the model integrates the video stream to $\sigma=0$, writes the resulting plan, which is not decoded to pixels during control, into a KV cache, and then denoises the action chunk conditioned on it,
\begin{equation}
Z_k=\Phi^Z_\theta(\epsilon_Z\mid H_k),\qquad A_k=\Phi^A_\theta(\epsilon_A\mid Z_k,H_k).
\label{eq:wam_policy}
\end{equation}
With $\epsilon_Z,\epsilon_A\sim\mathcal{N}(0,I)$, \cref{eq:wam_policy} defines the WAM policy $\pi_\theta(A_k\mid H_k)$.
The cost of a chunk grows with both $N_v$ and $N_a$.

\paragraph{Autoregressive inference.}
The model rolls forward chunk by chunk with this cache.
The plan of a chunk stays in the cache only until the chunk is executed, after which it is replaced by the latent of the real observation and the executed actions.

\subsection{Few-Step Distillation of WAMs}
\label{sec:prelim_distill}

Few-step distillation trains a student to reproduce a high-NFE teacher with far fewer function evaluations~\cite{salimans2022progressive,song2023consistency}.
In our setting, teacher and student are WAM policies with the same architecture, both initialized from the parameters $\theta_0$ of the high-NFE policy to be accelerated.
The teacher $\pi_{\mathrm{T}}$ keeps $\theta_{\mathrm{T}}=\theta_0$ frozen together with its original high-NFE sampler.
The student $\pi_{\mathrm{S}}$ uses $\theta_{\mathrm{S}}=\theta_0+\Delta$, where the LoRA update~\cite{hu2022lora} $\Delta$ is the only trained parameter, and samples with $N_v/N_a=1/2$.
We write $\Phi^Z_{\mathrm{T}},\Phi^A_{\mathrm{T}}$ and $\Phi^Z_{\mathrm{S}},\Phi^A_{\mathrm{S}}$ for their samplers.
With a single step, the student plan reduces to one evaluation of the video stream,
\begin{equation}
\Phi^Z_{\mathrm{S}}(\epsilon_Z\mid H)=\epsilon_Z-v^Z_{\theta_{\mathrm{S}}}(\epsilon_Z,1\mid H),
\label{eq:student_onestep}
\end{equation}
whereas the teacher runs its full high-NFE sampler.

\paragraph{Closed-loop exposure bias.}
A WAM policy acts in closed loop: at chunk $k$ it samples an action chunk, the chunk is executed, and the context advances,
\begin{equation}
A_k\sim\pi(\cdot\mid H_k),\qquad H_{k+1}=(\ell,\,z_{\le k+1},\,A_{\le k}),
\label{eq:closed_loop}
\end{equation}
where $z_{k+1}$ is the latent of the observation reached after executing $A_k$.
The distribution $d_\pi$ of contexts that $\pi$ visits therefore depends on $\pi$ itself.
A student trained only on contexts from demonstrations or from the teacher meets other contexts at deployment once its errors change its actions, and these errors can compound over the following chunks; training on the learner's own states counters this exposure bias~\cite{ross2011dagger,bengio2015scheduled}.
A few-step student adds a second mismatch within each inference: it samples through its own truncated chain, and in a WAM its action stream is conditioned on its own plan rather than on the teacher's.
On-policy distillation addresses such mismatches by supervising the student on its own samples~\cite{agarwal2024gkd}, and Ni~\etal~\cite{ni2026multiteacher} cover both the environment and the denoising level for flow-based VLAs, with several teachers and without reducing steps.
Training contexts can be drawn from a mixture of the student's and the teacher's closed-loop distributions, $\bar d=(1-\eta)\,d_{\pi_{\mathrm{S}}}+\eta\,d_{\pi_{\mathrm{T}}}$, which \cref{eq:objective} uses for OPSD (\cref{sec:opd}).

\section{Method Details}
\label{sec:suppl_method}

\subsection{OPSD Procedure}
\label{sec:suppl_opsd}

\Cref{alg:opd} summarizes OPSD (\cref{sec:opd}).

\begin{algorithm}[t]
\caption{On-Policy Self-Distillation (OPSD)}
\label{alg:opd}
\begin{algorithmic}[1]
\Require teacher $\pi_{\mathrm{T}}$, student $\pi_{\mathrm{S}}$ with $\theta_{\mathrm{S}}=\theta_0+\Delta$, mixture weight $\eta$
\State Run $\pi_{\mathrm{T}}$ in closed loop; cache each context with its $(\epsilon_Z,\epsilon_A,Z_{\mathrm{T}},A_{\mathrm{T}})$
\State $\Delta\gets 0$
\Repeat
  \State Run $\pi_{\mathrm{S}}$ in closed loop; collect contexts $H$ at its chunk boundaries \Comment{skipped if $\eta=1$}
  \State For each new $H$, draw $(\epsilon_Z,\epsilon_A)$ and label $(Z_{\mathrm{T}},A_{\mathrm{T}})$ by \cref{eq:pairs}
  \State Form a batch with a fraction $\eta$ from the cache and the rest from the new contexts
  \State Compute $\mathcal{L}$ by \cref{eq:loss}; update $\Delta$ and its EMA $\bar\Delta$
  \State Synchronize the student weights $\theta_0+\Delta$ to the environments
\Until{the training budget is reached}
\State \Return $\pi_{\mathrm{S}}$ with $\theta_{\mathrm{S}}=\theta_0+\bar\Delta$
\end{algorithmic}
\end{algorithm}

\subsection{Task-Specific Reinforcement Learning}
\label{sec:suppl_rl}

This section details the reinforcement learning of Stage~2.1 (\cref{sec:rl}), which follows DiffusionNFT~\cite{zheng2026diffusionnft}: a flow model is reinforced through a flow-matching loss on its own samples, without likelihood estimation or a critic.

\paragraph{Candidate windows.}
Starts are states at chunk boundaries of the OPSD teacher's own closed-loop rollouts, collected once before training.
In each round, the environment is restored to a start with context $H$, and the sampling policy $\pi_{\mathrm{old}}$, a slowly updated copy of the trained policy, runs $J=8$ candidate windows of $L$ chunks from it in closed loop; a round uses 64 starts, and a chunk holds 16 commands, which cover $0.32$\,s.
Candidate $j$ draws its own noise pair $(\epsilon^{j}_Z,\epsilon^{j}_A)$ and reuses it in every chunk,
\begin{equation}
Z^{j}_{k}=\Phi^Z_{\mathrm{old}}(\epsilon^{j}_Z\mid H^{j}_{k}),\quad
A^{j}_{k}=\Phi^A_{\mathrm{old}}(\epsilon^{j}_A\mid Z^{j}_{k},H^{j}_{k}),
\label{eq:window}
\end{equation}
where $k=1,\dots,L$ counts the chunks of the window, $H^{j}_{1}=H$, and $H^{j}_{k+1}$ follows from executing $A^{j}_{k}$ as in \cref{eq:closed_loop}.
The candidates of a start thus differ only in their noise, which they keep over the window so that their differences do not cancel from chunk to chunk.

\paragraph{Window rewards and weights.}
Each candidate receives a single reward $R^{j}$ for its window, not one per step, computed only from the trajectory of the task object or of the foot relative to it; the two rewards are defined below.
Rewards are not comparable across starts, so a candidate is compared only with the other candidates of its start,
\begin{equation}
w^{j}=\frac12+\frac12\,\operatorname{clip}\!\Big(\frac{R^{j}-\bar R}{\kappa},-1,1\Big),\quad
\bar R=\frac1J\sum_{j=1}^{J}R^{j}.
\label{eq:rl_weight}
\end{equation}
A weight near $1$ marks a candidate to move toward and a weight near $0$ one to move away from.
The scale $\kappa$ is the pooled standard deviation of the rewards within the starts of the first round,
\begin{equation}
\kappa^2=\frac{\sum_{n}\sum_{j}\big(R^{j}_{n}-\bar R_{n}\big)^2}{\sum_{n}(J-1)},
\label{eq:rl_scale}
\end{equation}
where $n$ runs over these starts; it is then kept fixed, so that the weights do not amplify noise once the candidates of a start differ less.
A start is skipped when the range of its rewards, $\max_j R^{j}-\min_j R^{j}$, is below $4$\,mm for the box and $0.5$\,cm for the door.

\paragraph{Reward for pushing the box.}
Starts are chunk boundaries after the foot first touches the box, and a window has $L=6$ chunks ($1.92$\,s), all of which give training samples.
The box is to be pushed along the $+x$ axis of the world frame into the target area.
Let $t_{\mathrm{s}}$ be the time $2.0$\,s after reset, at which the box has settled, $p_i(t)$, $i=1,\dots,4$, the bottom corners of the box, and $s(t)$ the displacement of its center along $+x$ since $t_{\mathrm{s}}$.
The reference pose translates the settled box along $+x$ without rotating it, and the deviation $e(t)$ is the mean distance of the corners to this pose,
\begin{equation}
\begin{aligned}
p^{\mathrm{ref}}_i(t)&=p_i(t_{\mathrm{s}})+\operatorname{clip}\big(s(t),0,u_{\mathrm{stop}}\big)\,\hat{x},\\
e(t)&=\frac14\sum_{i=1}^{4}\big\lVert p_i(t)-p^{\mathrm{ref}}_i(t)\big\rVert,
\end{aligned}
\label{eq:rl_push_dev}
\end{equation}
where $\hat{x}$ is the unit vector along $+x$ and $u_{\mathrm{stop}}$ is the advance at which the rearmost bottom corner reaches the near edge of the target area.
A lateral shift, a rotation, or a tilt of the box increases $e$, whereas its advance along $+x$ does not.
For a window from $t_0$ to $t_1=t_0+1.92$\,s, the reward is
\begin{equation}
R=-\big(e(t_1)-e(t_0)\big)+\tfrac12\min\big(s(t_1)-s(t_0),\,c\big).
\label{eq:rl_push_reward}
\end{equation}
The cap $c$ is the smaller of two advances: that of the median progress curve of 80 runs of the procedure that synthesizes the demonstrations of the task, over $1.92$\,s from the point where this curve reaches $s(t_0)$, and the remaining advance $u_{\mathrm{stop}}-s(t_0)$.
Advancing faster than these runs therefore adds nothing to the reward.
When the task succeeds inside the window, the final state is held until $t_1$.

\paragraph{Reward for opening the door.}
Each rollout of the OPSD teacher gives one start, the chunk boundary 4 chunks before the chunk in which the foot first makes contact with the handle, so that the candidates can still place the foot differently.
A window has $L=13$ chunks ($4.16$\,s), of which only the first 7 give training samples; the later chunks only serve the reward.
In the frame of the closed door, let $(d_x,d_y,d_z)$ be the position of the front-left foot relative to a reference point on the handle, with $d_x$ positive toward the door and $d_z$ vertical.
The contact time $t_{\mathrm{c}}$ is the first time, after the foot has risen above this point, at which
\begin{equation}
\sqrt{d_x^2+d_z^2}\le 3.4\,\mathrm{cm},\quad -6.5\,\mathrm{cm}\le d_y\le 7.0\,\mathrm{cm}.
\label{eq:rl_door_contact}
\end{equation}
The reward is how far toward the door the foot rests after contact, up to a cap,
\begin{equation}
R=\min\big(X,\,0.3\,\mathrm{cm}\big),
\label{eq:rl_door_reward}
\end{equation}
where $X$ is the mean of $d_x$ from $1.28$\,s to $1.92$\,s after $t_{\mathrm{c}}$.
The cap keeps a foot that is already far enough toward the door from being rewarded for pressing farther.
A candidate for which $X$ is undefined, because the foot makes no contact or the window ends before this interval does, receives the lowest reward among the candidates of its start, and a start without any defined $X$ is skipped.

\paragraph{Update.}
Training samples $y$ are the plans and action chunks of \cref{eq:window}, taken from the whole window or, when its later chunks only serve the reward, from its first chunks.
Each sample keeps its context, with the action conditioned on its plan, and carries the weight $w$ of its candidate.
With $y$ noised to $y_\sigma$, the velocity $v_\theta$ of the trained policy, $\theta=\theta_0+\Delta_g$, is contrasted with the velocity $v_{\mathrm{old}}$ of the sampling policy through two implicit branches,
\begin{equation}
\begin{aligned}
v^{\pm}&=v_{\mathrm{old}}\pm\beta\,(v_\theta-v_{\mathrm{old}}),\\
\mathcal{L}_{\mathrm{RL}}&=\tfrac{1}{\beta}\big[w\,\mathcal{E}(v^{+})+(1-w)\,\mathcal{E}(v^{-})\big]+\lambda_{\mathrm{ref}}\lVert v_\theta-v_{\theta_0}\rVert^2,
\end{aligned}
\label{eq:rl_loss}
\end{equation}
where $\mathcal{E}(v)$ is the flow-matching regression error of $v$ toward $\epsilon-y$ with the self-normalized weighting of DiffusionNFT, and $\mathcal{L}_{\mathrm{RL}}$ is summed over the two streams.
$\mathcal{E}(v^{+})$ decreases when $v_\theta$ moves from $v_{\mathrm{old}}$ toward the sample and $\mathcal{E}(v^{-})$ when it moves away, so $w$ sets the direction in which a candidate pulls the policy and $\beta$ limits the step.
The last term regularizes the specialist toward the OPSD teacher.
After each round, the sampling policy follows by $\theta_{\mathrm{old}}\leftarrow\mu\,\theta_{\mathrm{old}}+(1-\mu)\,\theta$.

\begin{figure}[t]
  \centering
  \includegraphics[width=\linewidth]{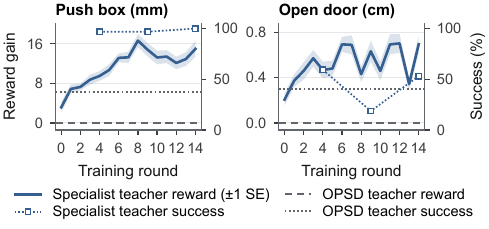}
  \caption{\textbf{Specialist teachers during Stage~2.1.}
  Solid line (left axis): reward gain over the OPSD teacher on held-out start states after each round, paired by start state and noise, with a band of $\pm1$ standard error over scenes; the two rewards measure different quantities (the box reward of \cref{eq:rl_push_reward} in mm, the offset of the foot on the door handle in cm).
  Squares (right axis): closed-loop success over 32 episodes after rounds 4, 9, and 14; the dotted line is the success of the OPSD teacher.
  For opening the door, the success falls from 19 of 32 episodes after round 4 to 6 after round 9 while the reward gain stays positive, so the reward is only a proxy of success there; the specialist teacher of \cref{tab:qlm} is the update after round 4 (\cref{sec:suppl_impl}).}
  \label{fig:rl_reward}
\end{figure}

\Cref{fig:rl_reward} shows the reward gain of the two specialist teachers over the OPSD teacher during training, with their closed-loop success.

\begin{table*}[t]
  \centering
  \footnotesize
  \caption{\textbf{Ablations on QLM-Bench Standard100 per task} (success rate, \%), with all rows of the ablation tables of the main paper.
  The protocol, $\pm$, and Seeds are as in \cref{tab:qlm}, and Updates counts the updates of the distillation.
  Groups marked with policy seed 42 use that seed only (100 episodes, 10 per task), so they have no $\pm$ and one entry of Seeds.}
  \label{tab:suppl_abl_full}
  \setlength{\tabcolsep}{2.5pt}
  \begin{tabular}{lcc*{10}{c}cc}
    \toprule
    Variant & NFE & Updates & Fridge & Oven & Valve & Drawer & Door & Press & Basket & Kick & Push & Sweep & All & Seeds \\
    \midrule
    \multicolumn{15}{l}{\emph{Training contexts (\cref{tab:abl_contexts})}} \\
    Teacher cache only ($\eta=1$) & 1/2 & 300 & 100 & 100 & 100 & 100 & 33 & 90 & 100 & 100 & 47 & 0 & 77.0$\pm$3.5 & 79/73/79 \\
    Streaming only ($\eta=0$) & 1/2 & 300 & 100 & 100 & 100 & 97 & 50 & 93 & 100 & 100 & 23 & 0 & 76.3$\pm$4.0 & 80/72/77 \\
    \rowcolor{oursrow}
    \textbf{OPSD (Ours, $\eta=0.5$)} & 1/2 & 300 & 100 & 100 & 100 & 97 & 37 & 83 & 100 & 100 & 50 & 0 & 76.7$\pm$4.7 & 82/75/73 \\
    \midrule
    \multicolumn{15}{l}{\emph{Longer training, policy seed 42 (\cref{tab:abl_longer})}} \\
    Teacher cache only ($\eta=1$) & 1/2 & 300 & 100 & 100 & 100 & 100 & 60 & 90 & 100 & 100 & 40 & 0 & 79.0 & 79 \\
    Teacher cache only ($\eta=1$) & 1/2 & 2,000 & 100 & 100 & 100 & 80 & 90 & 80 & 100 & 90 & 30 & 0 & 77.0 & 77 \\
    \rowcolor{oursrow}
    \textbf{OPSD (Ours, $\eta=0.5$)} & 1/2 & 300 & 100 & 100 & 100 & 100 & 80 & 90 & 100 & 100 & 50 & 0 & 82.0 & 82 \\
    \rowcolor{oursrow}
    \textbf{OPSD (Ours, $\eta=0.5$)} & 1/2 & 900 & 100 & 100 & 100 & 90 & 90 & 70 & 100 & 100 & 40 & 0 & 79.0 & 79 \\
    \midrule
    \multicolumn{15}{l}{\emph{Distillation objective, policy seed 42 (\cref{tab:abl_objective})}} \\
    Student w/o distillation & 1/2 & 0 & 100 & 100 & 50 & 0 & 90 & 100 & 100 & 90 & 50 & 0 & 68.0 & 68 \\
    Distribution matching ($\eta=0$) & 1/2 & 900 & 100 & 100 & 0 & 40 & 80 & 60 & 90 & 90 & 30 & 0 & 59.0 & 59 \\
    \rowcolor{oursrow}
    \textbf{OPSD (Ours, $\eta=0.5$)} & 1/2 & 900 & 100 & 100 & 100 & 90 & 90 & 70 & 100 & 100 & 40 & 0 & 79.0 & 79 \\
    \midrule
    \multicolumn{15}{l}{\emph{Sampling steps without distillation (\cref{tab:abl_steps})}} \\
    Teacher & 25/50 & 0 & 100 & 100 & 100 & 97 & 53 & 77 & 100 & 97 & 47 & 0 & 77.0$\pm$4.0 & 77/73/81 \\
    W/o distillation (2/4-NFE) & 2/4 & 0 & 100 & 100 & 67 & 97 & 53 & 100 & 100 & 100 & 63 & 0 & 78.0$\pm$1.7 & 79/79/76 \\
    Student w/o distillation & 1/2 & 0 & 100 & 100 & 47 & 0 & 93 & 93 & 100 & 97 & 60 & 3 & 69.3$\pm$1.2 & 68/70/70 \\
    \midrule
    \multicolumn{15}{l}{\emph{Proprioception in Stage~1.1, policy seed 42 (\cref{tab:abl_proprio})}} \\
    RGB + proprioception & 25/50 & 0 & 100 & 100 & 0 & 100 & 0 & 70 & 100 & 100 & 100 & 0 & 67.0 & 67 \\
    RGB only (Teacher) & 25/50 & 0 & 100 & 100 & 100 & 90 & 100 & 70 & 100 & 100 & 10 & 0 & 77.0 & 77 \\
    \bottomrule
  \end{tabular}
\end{table*}
\section{Details of the QLM-Bench Ablations}
\label{sec:suppl_qlm_ablation}

\begin{figure}[t]
  \centering
  \includegraphics[width=\linewidth]{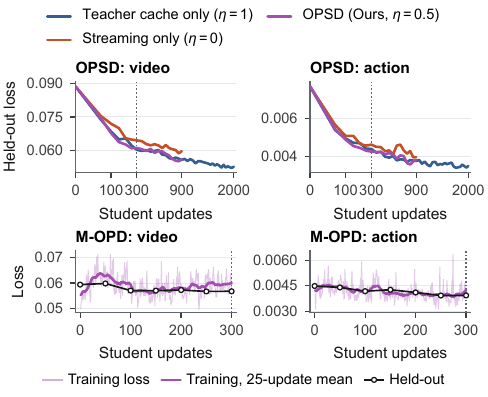}
  \caption{\textbf{Distillation loss.} The held-out loss is evaluated every 50 updates.
  Top: held-out loss of the students of the three training-context sources of \cref{tab:abl_contexts}, on the same 64 held-out contexts, with a square-root horizontal axis; the dotted line marks 300 updates, after which the students are evaluated.
  The held-out contexts are drawn from the teacher cache, so the curves do not rank the sources: they disadvantage Streaming only, which trains on no cached contexts.
  Bottom: the M-OPD student (\cref{sec:mopd}), which starts from the OPSD student and learns from the specialist teachers routed by task.
  The light line is the training loss of every update, computed on the contexts sampled for that update rather than on the held-out contexts; the dark line is its centered moving average over 25 updates, with shorter windows at both ends; the circles are the held-out loss.
  The held-out contexts of M-OPD come from its own teacher cache, so its loss is not on the scale of the top panels.}
  \label{fig:opsd_loss}
\end{figure}

\Cref{tab:suppl_abl_full} gives the ablations of \cref{sec:exp_ablation} per task.
Groups marked with policy seed 42 are single-seed results.
The RGB-only (Teacher) row of \cref{tab:abl_proprio} is the Teacher of \cref{tab:qlm} at policy seed 42 (77 of 100 episodes, the first entry of its Seeds).
Per-task rates at a single seed can differ widely from the three-seed rates: with seed 42 the Teacher opens the door in 10 of 10 episodes and pushes the box in 1 of 10, against 16 of 30 and 14 of 30 over the three seeds.
\Cref{fig:opsd_loss} shows the held-out distillation loss of the three training-context sources of \cref{tab:abl_contexts} and of the M-OPD student.
The rest of this section specifies the two variants that are not defined by $\eta$.

\paragraph{Distribution-matching variant.}
It shares with OPSD the teacher, the 1/2-NFE sampler, the rank of $\Delta$, the optimizer (AdamW with learning rate $3\times10^{-5}$ after 20 warm-up updates, weight decay 0.01, and gradient clipping at 1.0), the EMA of the student weights (decay 0.99), the 24 streaming contexts per update, and the 900 updates.
It differs in four settings.
(i) The loss follows DMD~\cite{yin2024dmd,yin2024dmd2}: the update of the student is the difference between the scores of the frozen teacher and of a fake model, normalized per context and per stream.
(ii) The fake model is a second LoRA of the same rank on $\theta_0$, trained on the student's outputs with five updates per student update, after five initial rounds that update only the fake model.
(iii) The action term has weight 0.2, whereas OPSD uses $\lambda_A=1$ in \cref{eq:loss}.
(iv) There is no teacher cache, so every update uses the 24 streaming contexts alone ($\eta=0$).
We evaluate its EMA weights after 900 updates.

\paragraph{Proprioception variant.}
The proprioceptive state has 10 dimensions per observed frame: the height of the base above the ground (1), the direction of gravity in the body frame (3), and the linear (3) and angular (3) velocity of the base in the body frame.
It has its own input embedding and output projection, is denoised jointly with the visual latents, and adds a flow-matching loss of weight 1 to the video and action losses; the action stream conditions on both the visual latents and this state.
Height and velocities are normalized with their 1st and 99th percentiles over the training demonstrations.
Everything else follows the teacher: the same pretrained LingBot-VA, the same 80 demonstrations per task, 5,000 updates with the same optimizer settings, the 9-dimensional command, and evaluation with 25 video and 50 action steps without classifier-free guidance.

\section{Generalization to an Arm WAM on LIBERO}
\label{sec:suppl_arm}

OPSD (Stage 1.2 of WAM-Policy) does not depend on the embodiment (\cref{sec:method}).
We therefore also apply it to the arm WAM LingBot-VA~\cite{li2026lingbotva} on LIBERO, and ablate its design there.

\subsection{Setup}
\label{sec:suppl_arm_setup}
For the arm WAM we use LIBERO-Long~\cite{liu2023libero}, with 50 episodes for each of its 10 tasks, and LIBERO-Pro2.0~\cite{zhou2026liberopro2}, which adds 22 static and 20 runtime perturbations.
On LIBERO-Pro2.0 we follow the official test split of LIBERO-Long, which holds out two tasks.
Two of the perturbations apply to only one of these tasks, so the split has 84 task instances and 2,520 episodes over 30 seeds.
We report the official group success rates for the unperturbed (BASE), static, and runtime settings.
The two benchmarks cover different tasks and episodes, so we do not compare their success rates with each other.
The Stage~1 policies of \cref{sec:exp_setup} are built in the same way from LingBot-VA, and \cref{sec:suppl_arm_stats} lists their paired comparisons.
Here $\theta_0$ is the official LingBot-VA checkpoint for LIBERO-Long, and the teacher samples with 20 video and 50 action steps and guidance scales of 5.0 for video and 1.0 for action.
The loss, the student sampler, the LoRA rank, and the mixture weights are those of the quadruped (\cref{sec:exp_setup}).

\subsection{Results}
\label{sec:suppl_arm_results}
\begin{table}[t]
  \centering
  \footnotesize
  \caption{\textbf{Success rate (\%) of an arm WAM on LIBERO.}
  LIBERO-Pro2.0 reports the official group rates; BASE has 60 episodes.
  Every student is evaluated after 300 updates, and Streaming only is not evaluated on LIBERO-Long (--).
  $^\dagger$Published numbers under different protocols, not run by us: LIBERO-Long from Li~\etal~\cite{li2026lingbotva}, and static and runtime values are our means over the perturbations in the table of Zhou~\etal~\cite{zhou2026liberopro2}, which lists no BASE rate for these tasks.}
  \label{tab:libero}
  \setlength{\tabcolsep}{3pt}
  \begin{tabular}{lcccc}
    \toprule
    & \multicolumn{3}{c}{LIBERO-Pro2.0} & \\
    \cmidrule(lr){2-4}
    Policy & BASE & Static & Runtime & Long \\
    \midrule
    Teacher (high-NFE)$^\dagger$ & -- & 63.7 & 79.6 & 98.5 \\
    Student w/o distillation & 95.0 & 64.4 & 68.5 & 91.8 \\
    \midrule
    Teacher cache only ($\eta=1$) & 100.0 & 69.2 & 70.8 & 96.4 \\
    Streaming only ($\eta=0$) & 93.3 & 70.2 & 72.4 & -- \\
    \rowcolor{oursrow}
    \textbf{OPSD (Ours)} & 98.3 & 68.0 & 72.3 & 98.2 \\
    \bottomrule
  \end{tabular}
\end{table}

We apply the distillation to LingBot-VA on LIBERO, with the same loss, student sampler, and LoRA rank as on the quadruped (\cref{tab:libero}).
On LIBERO-Long, OPSD raises the success of the student without distillation from 91.8\% to 98.2\%, a gain of 6.4 points.
It is also 1.8 points above the teacher-cache-only variant.
On LIBERO-Pro2.0, OPSD gains 3.6 points under static and 3.8 points under runtime perturbations over the student without distillation (overall 3.9 points, the task-averaged difference of \cref{tab:paired_libero}).
The streaming-only variant gains 5.8 and 3.9 points (overall 4.9 points), 1.0 points above OPSD overall.
The teacher-cache-only variant gains 4.8 and 2.3 points (overall 3.8 points).
Compared with the teacher-cache-only variant, OPSD is 1.1 points lower under static and 1.5 points higher under runtime perturbations, with an overall difference of 0.1 points, and the streaming-only variant is 1.1 points higher overall; differences are computed from unrounded rates, so they can deviate by 0.1 from those of the rounded entries of \cref{tab:libero}.
\Cref{tab:paired_libero} gives the confidence intervals and McNemar tests of these comparisons.
Distillation thus improves a few-step arm WAM on both nominal and perturbed suites, which supports that OPSD does not depend on the embodiment.

\subsection{Ablations}
\label{sec:suppl_arm_ablation}
\begin{table}[t]
  \centering
  \footnotesize
  \caption{\textbf{Ablation studies on LIBERO-Pro2.0: success rate (\%).}
  SFT only receives 692 updates, whose supervised chunks match 2,000 updates of the teacher-cache-only variant, 6.7 times the budget of the distilled students.}
  \label{tab:ablation}
  \setlength{\tabcolsep}{2.5pt}
  \begin{tabular}{lcccc}
    \toprule
    Variant & Updates & BASE & Static & Runtime \\
    \midrule
    Teacher cache only ($\eta=1$) & 300 & 100.0 & 69.2 & 70.8 \\
    Streaming only ($\eta=0$) & 300 & 93.3 & 70.2 & 72.4 \\
    \rowcolor{oursrow}
    \textbf{OPSD (Ours)} & 300 & 98.3 & 68.0 & 72.3 \\
    \midrule
    Collected once & 300 & 96.7 & 70.2 & 72.3 \\
    SFT only (no teacher) & 692 & 90.0 & 65.9 & 69.7 \\
    \bottomrule
  \end{tabular}
\end{table}

We ablate two choices on LIBERO-Pro2.0 against the students of \cref{tab:libero} (\cref{tab:ablation}).

\paragraph{Repeated collection.}
The collected-once variant collects its contexts once with the initial student, about as many as the teacher cache, and trains only on this fixed set, as in the one-time collection of Yang~\etal~\cite{yang2026wamopd}.
Like the streaming-only variant, it uses no teacher cache, so comparing the two tests whether the contexts must follow the student as it changes.
It draws 48 contexts per update from its fixed set, whereas the streaming-only variant draws 24 new ones.
It reaches 70.2\% and 72.3\% under static and runtime perturbations, nearly the same as the 70.2\% and 72.4\% of the streaming-only variant, and the two differ by 0.1 points overall (\cref{tab:paired_libero}).
Overall, it is 5.0 points above the student without distillation, 1.2 points above the teacher-cache-only variant, and 1.1 points above OPSD (\cref{tab:paired_libero}).
Across OPSD and its three variants, overall success on LIBERO-Pro2.0 differs by at most 1.2 points: as on QLM-Bench (\cref{sec:exp_ablation}), the gain comes from the shared distillation objective, whatever the source of the training contexts.

\paragraph{Teacher.}
SFT only (no teacher) fine-tunes a LoRA of the same rank on the LingBot-VA LIBERO demonstrations with the official flow-matching recipe, and samples with 1/2 NFE.
Despite its larger budget, it reaches only 65.9\% and 69.7\%, below the distilled students, so the gain comes from matching the teacher rather than from further fine-tuning.
Overall, it is 2.6 points below the teacher-cache-only variant and 2.7 points below OPSD, and only 1.2 points above the student without distillation (\cref{tab:paired_libero}).

\subsection{Longer Training}
\label{sec:suppl_longer}

\begin{table}[t]
  \centering
  \footnotesize
  \caption{\textbf{Success rate (\%) on LIBERO after longer training.}
  The 300-update rows are those of \cref{tab:libero}.
  \Cref{tab:abl_longer} gives the corresponding QLM-Bench results.}
  \label{tab:longer}
  \setlength{\tabcolsep}{2pt}
  \begin{tabular}{lccccc}
    \toprule
    & & \multicolumn{3}{c}{LIBERO-Pro2.0} & \\
    \cmidrule(lr){3-5}
    Policy & Updates & BASE & Static & Runtime & Long \\
    \midrule
    Teacher cache only ($\eta=1$) & 300 & 100.0 & 69.2 & 70.8 & 96.4 \\
    Teacher cache only ($\eta=1$) & 2,000 & 100.0 & 68.9 & 73.5 & 96.8 \\
    \midrule
    \rowcolor{oursrow}
    \textbf{OPSD (Ours)} & 300 & 98.3 & 68.0 & 72.3 & 98.2 \\
    \rowcolor{oursrow}
    \textbf{OPSD (Ours)} & 900 & 95.0 & 69.5 & 72.1 & 97.2 \\
    \bottomrule
  \end{tabular}
\end{table}

\Cref{tab:longer} reports the teacher-cache-only variant after 2,000 updates and OPSD after 900 updates, next to the students after 300 updates.
On LIBERO-Pro2.0, the overall difference of the 300-update student from its longer-trained counterpart is $-1.1$ points for the teacher-cache-only variant and $-0.6$ points for OPSD, and OPSD after 900 updates differs from the teacher-cache-only variant after 2,000 updates by $-0.4$ points.
\Cref{tab:paired_libero} gives the corresponding tests.

\begin{figure*}[t]
  \centering
  \includegraphics[width=\linewidth]{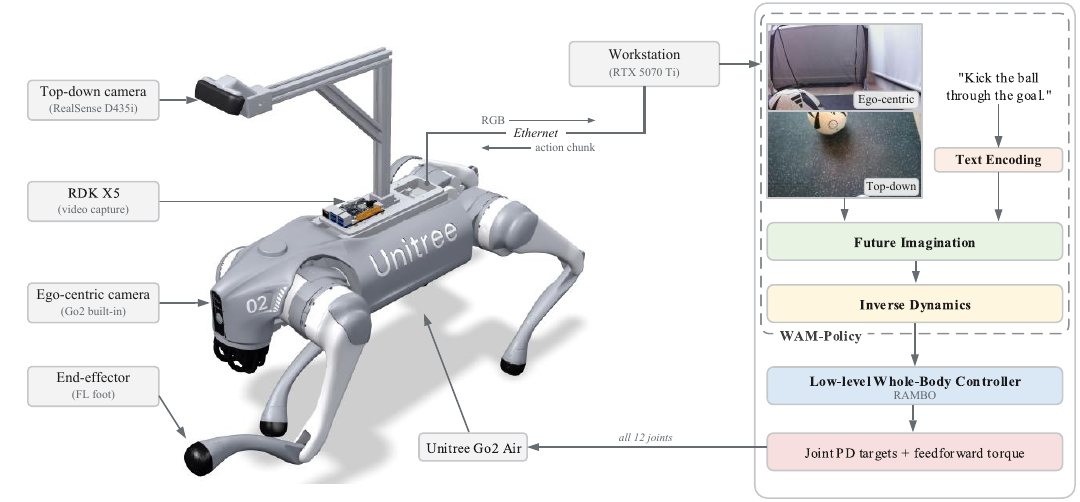}
  \caption{\textbf{Real-robot setup} of \cref{sec:exp_real}.
  Left: the Unitree Go2 Air, which carries the two cameras and an RDK X5 board and manipulates with its front-left (FL) leg, connects to the workstation by Ethernet.
  Right: WAM-Policy takes the ego-centric and top-down views and the task prompt; the video stream predicts the plan, the latent of the next observation, and the action stream infers the action chunk from it by inverse dynamics. The whole-body controller RAMBO~\cite{cheng2025rambo} executes the chunk on the robot with an RL policy and joint PD control.}
  \label{fig:suppl_real_setup}
\end{figure*}

\section{Real-Robot Setup}
\label{sec:suppl_real}

\Cref{fig:suppl_real_setup} shows the deployment of \cref{sec:exp_real}.
The whole-body controller $\mathcal{C}$ is RAMBO~\cite{cheng2025rambo}, which executes each command of the chunk with an RL policy.
The supplementary videos show the real-robot rollouts.

\section{Implementation Details}
\label{sec:suppl_impl}

This section completes the implementation details of \cref{sec:exp_setup} for QLM-Bench.

\begin{figure}[t]
  \centering
  \includegraphics[width=\linewidth]{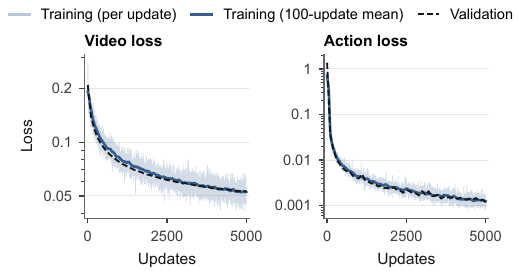}
  \caption{\textbf{Fine-tuning losses of the transferred policy} (Stage~1.1), on a logarithmic scale.
  Light: training loss of each update; dark: its centered moving average over 100 updates, \ie one pass over the 800 training demonstrations; black dashed: validation loss on 10 held-out demonstrations per task, every 100 updates.}
  \label{fig:suppl_finetune_loss}
\end{figure}

\paragraph{Fine-tuning.}
We fine-tune for 5,000 updates on 80 demonstrations per task and hold out the other 20.
The re-initialized input embedding and output projection of the action stream take the 9-dimensional command in place of the 30-dimensional dual-arm vector of LingBot-VA, which consists of end-effector poses, joint angles, and gripper commands.
\Cref{fig:suppl_finetune_loss} shows the losses of this fine-tuning: the validation loss falls from 0.209 to 0.053 for the video stream and from 1.37 to 0.0012 for the action stream, and most of each decrease, 72\% and 83\% of the drop in log loss, occurs within the first 1,000 updates.

\paragraph{Distillation.}
$\Delta$ has rank 256, and the teacher cache holds about 5,000 contexts.
OPSD draws 24 streaming and 24 cached contexts per update ($\eta=0.5$ in \cref{eq:objective}).
M-OPD starts from the EMA weights of the OPSD student after 300 updates and trains for another 300 updates with the settings of OPSD; the optimizer state, the warm-up, and the EMA start anew.
As in OPSD, streaming environments are balanced over tasks; specialist contexts replace the cache entries of their tasks one for one.
Neither the teacher nor the student uses classifier-free guidance, and every student is evaluated with EMA weights after 300 updates unless stated otherwise.

\paragraph{Hyperparameters.}
All steps use AdamW with a linear warm-up to a constant learning rate.
Stage~1.1 fine-tunes the transformer with a learning rate of $1\times10^{-5}$.
OPSD and M-OPD train the LoRA update with a learning rate of $3\times10^{-5}$, with $c=10^{-3}$ and $\lambda_A=1$ in \cref{eq:loss}, and keep an EMA of the trained weights with decay 0.99.
Each specialist of Stage~2.1 trains a LoRA update of rank 32 with a learning rate of $1\times10^{-4}$, for 15 rounds of 24 updates.
Each camera renders $640\times360$ RGB at 50\,Hz; the model takes every fourth frame, resized to $320\times180$ per view, and the two views are stacked vertically after the VAE.
A chunk holds one latent, which covers four such frames, and 16 commands at 50\,Hz.
A command consists of the planar base velocity (m/s) and yaw rate (rad/s), the target position of the front-left foot (m), and its target force (N); the velocity and position dimensions are normalized with their 1st and 99th percentiles over the training demonstrations.
All samplers use Euler steps on a uniform grid of noise levels with a flow shift of 5 for the video stream and 1 for the action stream, and attend to a window of 8 past video and action units.

\paragraph{Leg-Manip.}
The per-task planners of Leg-Manip use 4 to 80 successful demonstrations per task, only 4 for pushing the box and 5 for sweeping the block; opening the door has none. Over the nine tasks with demonstrations, its mean success is 68.6\%.

\paragraph{Specialist teachers.}
Each specialist of \cref{tab:qlm} is applied on all ten tasks.
The push-box specialist is the update after the last round of Stage~2.1, chosen without looking at the evaluation; the open-door specialist is the update after round 4, chosen on 32 held-out episodes of training, not on Standard100.
Outside its tuned task, the push-box specialist pulls the drawer open in 21 of 30 episodes against 29 for the OPSD teacher, and the open-door specialist sweeps the block in 16 of 30 against none.
Both differences remain when the OPSD teacher is evaluated again on the second workstation with batched inference (\cref{sec:suppl_compute}): it pulls the drawer open in 29 of 30 episodes and never sweeps the block.

\begin{table}[t]
  \centering
  \footnotesize
  \caption{\textbf{Paired comparisons on QLM-Bench.}
  $\Delta$ is the success rate of the first policy minus that of the second, in points; $+/-$ counts the pairs in which only the first or only the second policy succeeds; $p$ is the exact two-sided McNemar test on these counts.
  T: Teacher, the OPSD teacher; W/o: Student w/o distillation; OPSD: our student, the Student (Ours) of \cref{tab:qlm}; M-OPD: the M-OPD student of \cref{tab:qlm}; tc, str: Teacher cache only ($\eta=1$) and Streaming only ($\eta=0$).
  Students are evaluated after 300 updates unless a superscript gives the number of updates.
  228 of the 300 episodes of the M-OPD student run under the second evaluation configuration of \cref{sec:suppl_compute}; its 60 episodes of pushing the box and opening the door run under the configuration of the other rows.}
  \label{tab:paired}
  \setlength{\tabcolsep}{3pt}
  \begin{tabular}{lrrl}
    \toprule
    Comparison & $\Delta$ & $+/-$ & $p$ \\
    \midrule
    \multicolumn{4}{l}{\emph{300 episodes}} \\
    tc vs.\ W/o & $+7.7$ & 54/31 & 0.017 \\
    str vs.\ W/o & $+7.0$ & 49/28 & 0.022 \\
    OPSD vs.\ W/o & $+7.3$ & 53/31 & 0.021 \\
    tc vs.\ T & $0.0$ & 12/12 & 1.0 \\
    str vs.\ T & $-0.7$ & 11/13 & 0.84 \\
    OPSD vs.\ T & $-0.3$ & 11/12 & 1.0 \\
    OPSD vs.\ tc & $-0.3$ & 10/11 & 1.0 \\
    OPSD vs.\ str & $+0.3$ & 10/9 & 1.0 \\
    str vs.\ tc & $-0.7$ & 10/12 & 0.83 \\
    M-OPD vs.\ W/o & $+15.7$ & 59/12 & $1\times10^{-8}$ \\
    M-OPD vs.\ T & $+8.0$ & 30/6 & $7\times10^{-5}$ \\
    M-OPD vs.\ OPSD & $+8.3$ & 32/7 & $7\times10^{-5}$ \\
    \midrule
    \multicolumn{4}{l}{\emph{Policy seed 42, 100 episodes}} \\
    tc$^{2000}$ vs.\ W/o & $+9$ & 16/7 & 0.093 \\
    OPSD$^{900}$ vs.\ W/o & $+11$ & 17/6 & 0.035 \\
    OPSD$^{900}$ vs.\ tc$^{2000}$ & $+2$ & 4/2 & 0.69 \\
    \bottomrule
  \end{tabular}
\end{table}

\begin{table}[t]
  \centering
  \footnotesize
  \caption{\textbf{Paired comparisons on LIBERO.}
  $\Delta$, $+/-$, and $p$ are as in \cref{tab:paired}, and 95\% CI is a bootstrap interval of $\Delta$.
  W/o: Student w/o distillation; OPSD: the OPSD (Ours) of \cref{tab:libero}; tc, str, once: Teacher cache only ($\eta=1$), Streaming only ($\eta=0$), and Collected once; SFT: SFT only (no teacher).
  Students are evaluated after 300 updates unless a superscript gives the number of updates.}
  \label{tab:paired_libero}
  \setlength{\tabcolsep}{3pt}
  \begin{tabular}{lrrcl}
    \toprule
    Comparison & $\Delta$ & $+/-$ & 95\% CI & $p$ \\
    \midrule
    \multicolumn{5}{l}{\emph{LIBERO-Pro2.0, 2,520 episodes}} \\
    tc vs.\ W/o & $+3.8$ & 237/143 & [1.5, 6.0] & $2\times10^{-6}$ \\
    str vs.\ W/o & $+4.9$ & 249/127 & [2.9, 6.8] & $3\times10^{-10}$ \\
    OPSD vs.\ W/o & $+3.9$ & 232/136 & [1.7, 6.0] & $6\times10^{-7}$ \\
    OPSD vs.\ tc & $+0.1$ & 127/125 & [$-1.1$, 1.3] & 0.95 \\
    str vs.\ OPSD & $+1.0$ & 142/116 & [$-0.3$, 2.5] & 0.12 \\
    str vs.\ tc & $+1.1$ & 141/113 & [$-0.1$, 2.4] & 0.090 \\
    once vs.\ W/o & $+5.0$ & 244/120 & [3.0, 7.1] & $8\times10^{-11}$ \\
    once vs.\ tc & $+1.2$ & 137/107 & [$-0.4$, 3.0] & 0.063 \\
    once vs.\ OPSD & $+1.1$ & 144/116 & [$-0.6$, 3.0] & 0.094 \\
    once vs.\ str & $+0.1$ & 125/123 & [$-1.2$, 1.3] & 0.95 \\
    SFT$^{692}$ vs.\ W/o & $+1.2$ & 214/184 & [$-0.1$, 2.5] & 0.15 \\
    SFT$^{692}$ vs.\ tc & $-2.6$ & 137/201 & [$-4.7$, $-0.5$] & $6\times10^{-4}$ \\
    SFT$^{692}$ vs.\ OPSD & $-2.7$ & 139/205 & [$-4.7$, $-0.8$] & $4\times10^{-4}$ \\
    tc vs.\ tc$^{2000}$ & $-1.1$ & 98/126 & [$-2.7$, 0.5] & 0.071 \\
    OPSD vs.\ OPSD$^{900}$ & $-0.6$ & 105/120 & [$-2.2$, 0.9] & 0.35 \\
    OPSD$^{900}$ vs.\ tc$^{2000}$ & $-0.4$ & 97/108 & [$-2.1$, 1.4] & 0.49 \\
    \midrule
    \multicolumn{5}{l}{\emph{LIBERO-Long, 500 episodes}} \\
    tc vs.\ W/o & $+4.6$ & 37/14 & [1.8, 7.4] & 0.0018 \\
    OPSD vs.\ W/o & $+6.4$ & 37/5 & [4.0, 8.8] & $4\times10^{-7}$ \\
    OPSD vs.\ tc & $+1.8$ & 14/5 & [0.2, 3.6] & 0.064 \\
    OPSD$^{900}$ vs.\ tc$^{2000}$ & $+0.4$ & 11/9 & [$-1.4$, 2.2] & 0.82 \\
    \bottomrule
  \end{tabular}
\end{table}

\section{Compute Resources}
\label{sec:suppl_compute}

\paragraph{Training time.}
Each training run uses one node with eight GPUs.
The fine-tuning of Stage~1.1 takes about 14 hours, OPSD about 8 hours for the 300 updates of the student in \cref{tab:qlm}, the two specialist teachers of Stage~2.1 about 5 and 9 hours, and M-OPD about 5 hours, shorter than OPSD because the collection of its streaming contexts was optimized.

\paragraph{Timing.}
Time in \cref{tab:qlm} is measured on an RTX 5070 Ti and counts only the sampling of the video and action streams, not the encoding of the observations.
The times of the teacher and of the 1/2-NFE sampler are measured on the teacher's parameters, with the high-NFE sampler (25 video and 50 action steps) and with the 1/2-NFE sampler, and each is the mean duration of one inference call, which generates one chunk.
The students use the same 1/2-NFE sampler and are timed with their LoRA updates merged into the weights, which gives 0.159\,s per chunk.
The specialist teachers are timed with the high-NFE sampler and their LoRA updates applied without merging them into the weights, which gives 2.746\,s and 2.740\,s per chunk, against 2.226\,s for the OPSD teacher.

\paragraph{Evaluation configurations.}
The Teacher, Student w/o distillation, and OPSD student rows of \cref{tab:qlm} and all rows of the ablations are evaluated under one configuration, on a workstation with an RTX 5070 Ti and the VAE on the CPU.
Of the 300 episodes of the push-box specialist teacher, 60 run under this configuration, and the other 240 on a second workstation with the VAE on the GPU, 220 of them with batched inference.
Of those of the open-door specialist teacher, 53 run under this configuration, including all 30 of opening the door, and the others on the second workstation.
Of those of the M-OPD student, 72 run under this configuration, including all 30 of pushing the box and all 30 of opening the door, and the other 228 on the second workstation, where one resident model serves several concurrent clients with the VAE on the GPU.
The parameters and the sampler are the same in both configurations, but the arithmetic is not bit-identical, so outside their tuned tasks the per-task differences of these three rows from the OPSD teacher are indicative only.

\section{Limitations and Future Work}
\label{sec:suppl_limitations}

\begin{itemize}
\item \textbf{Task coverage of Stage~2.}
Stage~2.1 tunes only two tasks, each with a reward designed for its task object.
Extending it to more tasks calls for rewards that need less task-specific design.
\item \textbf{Scope of evaluation.}
All results use one quadruped, one whole-body controller, and one manipulating leg, and Standard100 varies the scenes of the ten training tasks; transfer to other legged robots and generalization to unseen tasks or instructions are not evaluated.
\item \textbf{End-to-end latency.}
On the real robot, sampling takes 0.159\,s per chunk, but the closed loop takes 0.66\,s with the encoding of the observations, about twice the 0.32\,s that a chunk covers, so the loop is not yet real time.
Accelerating the encoder is the next step.
\item \textbf{Simulation-to-real gap.}
WAM-Policy is trained only in simulation, and the OPSD student succeeds less often on the real robot than in simulation (\cref{tab:real}).
Adding real-robot data is a natural next step.
\end{itemize}

\section{Statistical Analysis}
\label{sec:suppl_stats}

\subsection{QLM-Bench}

\Cref{tab:paired} lists paired comparisons between the policies of the main paper on QLM-Bench.
Two episodes form a pair when they share the case and the policy seed, which gives 300 pairs per comparison (100 for policy seed 42 alone).
For each comparison we report the exact two-sided McNemar test on the discordant pairs.

\subsection{LIBERO}
\label{sec:suppl_arm_stats}

\Cref{tab:paired_libero} lists paired comparisons between the policies of \cref{sec:suppl_arm}.
Two episodes form a pair when they share the task instance and the seed, which gives 2,520 pairs per comparison on LIBERO-Pro2.0 and 500 on LIBERO-Long.
For each comparison we report the exact two-sided McNemar test on the discordant pairs and a bootstrap 95\% confidence interval of $\Delta$ that resamples whole task instances within each task.
On LIBERO-Pro2.0, $\Delta$ is the mean over the two held-out tasks of the difference in success rate.

\section{Additional Simulation Rollouts}
\label{sec:suppl_rollouts}

\Crefrange{fig:suppl_rollout_fridge}{fig:suppl_rollout_sweep} show four closed-loop rollouts for each task, in the order of \cref{tab:qlm}: episodes of the OPSD student after 900 updates with policy seed 42 (79 of 100 successes, \cref{tab:abl_longer}) on four different Standard100 cases.
Time runs from left to right; the rows show the ego-centric and top-down observations, the two views as predicted by the policy (Prediction), \ie its plan decoded by the frozen VAE for display only, and a third-person view that the policy does not see (Observer).
The supplementary videos show further closed-loop rollouts in simulation.
\clearpage
\onecolumn
\nolinenumbers %
\captionsetup[figure]{skip=4pt}
\noindent\begin{minipage}{\textwidth}
  \captionsetup{type=figure}
  \centering
  \makebox[\textwidth][c]{\includegraphics[width=1.1\textwidth]{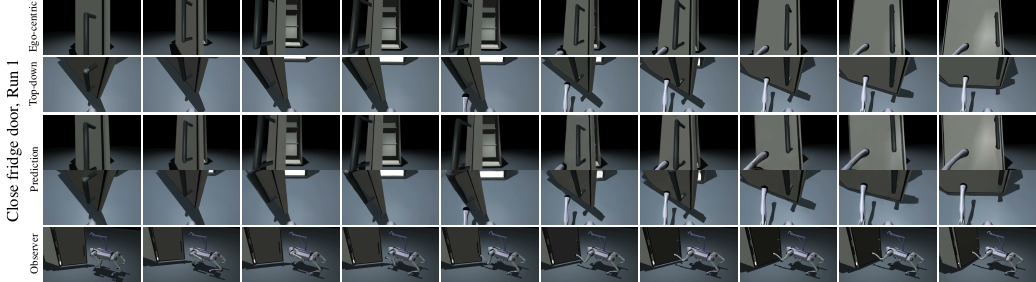}}\\[8pt]
  \makebox[\textwidth][c]{\includegraphics[width=1.1\textwidth]{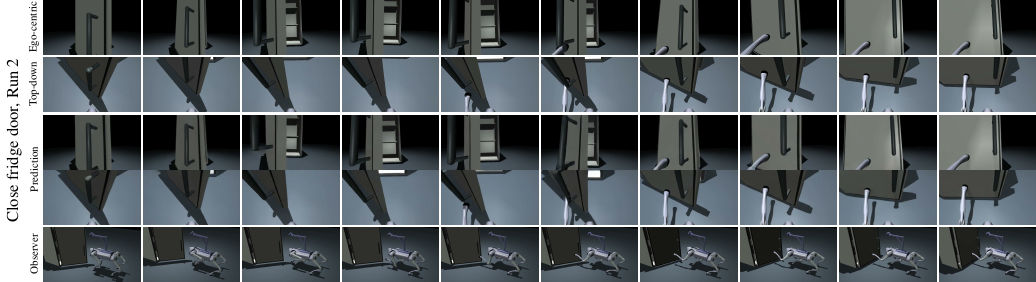}}\\[8pt]
  \makebox[\textwidth][c]{\includegraphics[width=1.1\textwidth]{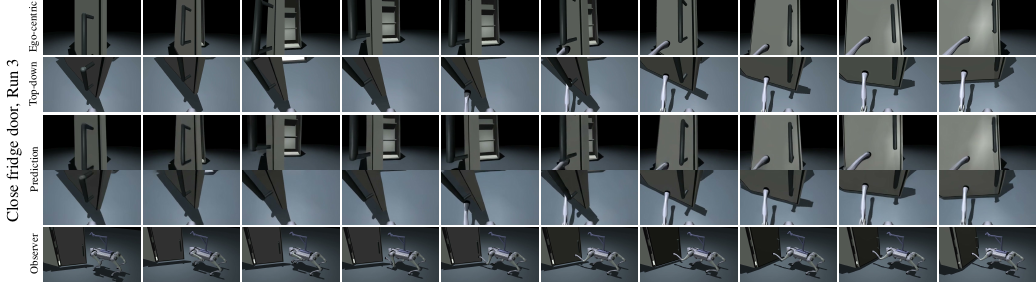}}\\[8pt]
  \makebox[\textwidth][c]{\includegraphics[width=1.1\textwidth]{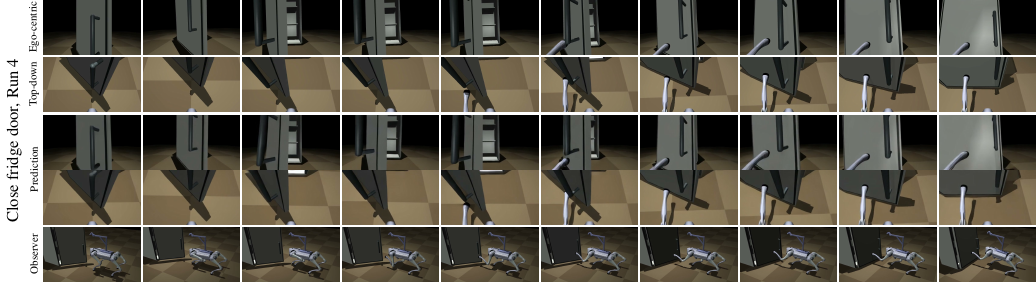}}
  \caption{\textbf{Closing the fridge door in QLM-Bench.} Runs 1 to 4 from top to bottom.}
  \label{fig:suppl_rollout_fridge}
\end{minipage}

\clearpage
\noindent\begin{minipage}{\textwidth}
  \captionsetup{type=figure}
  \centering
  \makebox[\textwidth][c]{\includegraphics[width=1.1\textwidth]{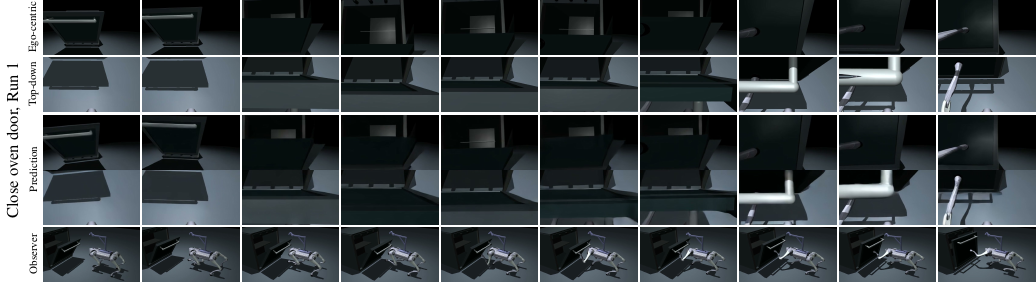}}\\[8pt]
  \makebox[\textwidth][c]{\includegraphics[width=1.1\textwidth]{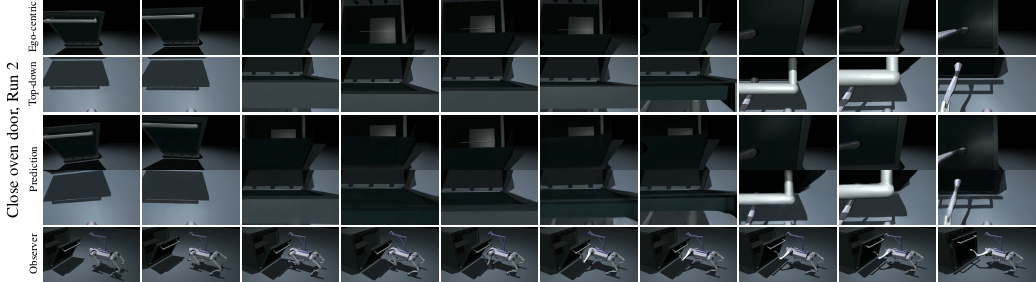}}\\[8pt]
  \makebox[\textwidth][c]{\includegraphics[width=1.1\textwidth]{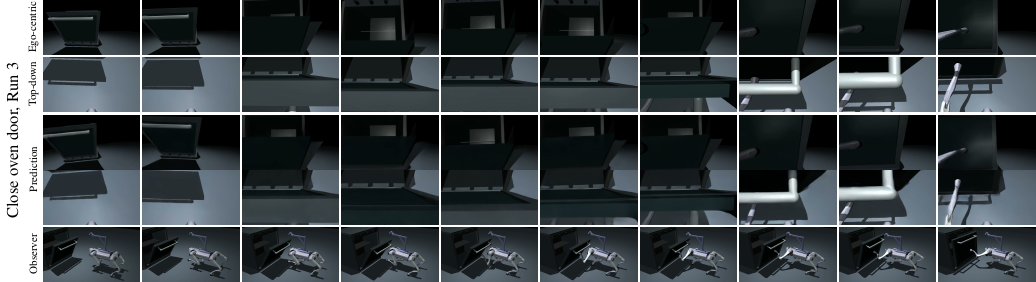}}\\[8pt]
  \makebox[\textwidth][c]{\includegraphics[width=1.1\textwidth]{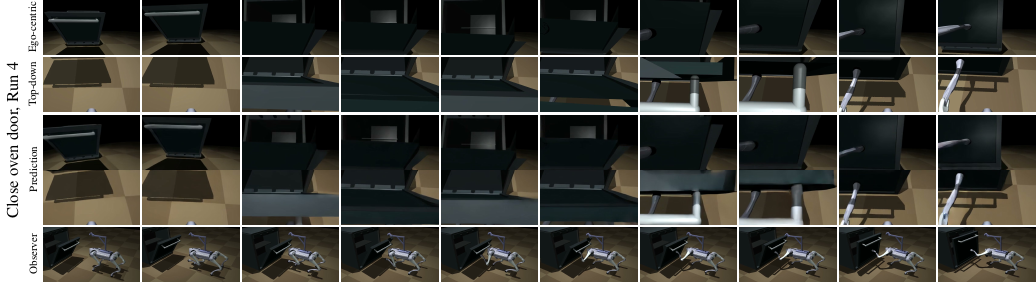}}
  \caption{\textbf{Closing the oven door in QLM-Bench.} Runs 1 to 4 from top to bottom.}
  \label{fig:suppl_rollout_oven}
\end{minipage}

\clearpage
\noindent\begin{minipage}{\textwidth}
  \captionsetup{type=figure}
  \centering
  \makebox[\textwidth][c]{\includegraphics[width=1.1\textwidth]{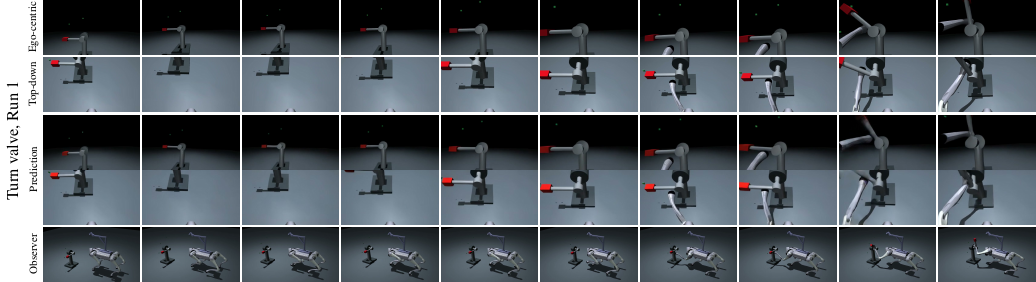}}\\[8pt]
  \makebox[\textwidth][c]{\includegraphics[width=1.1\textwidth]{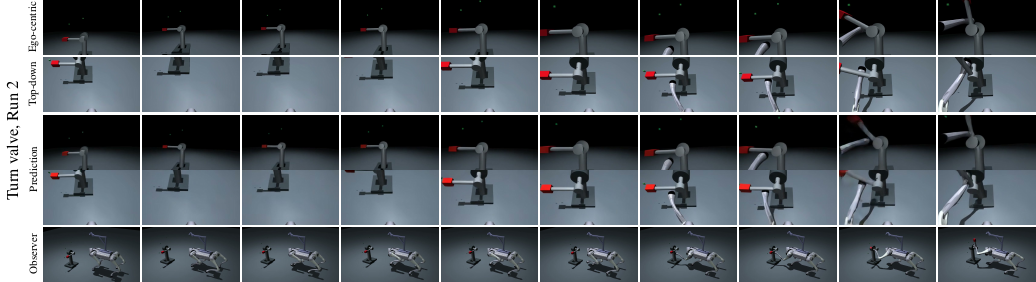}}\\[8pt]
  \makebox[\textwidth][c]{\includegraphics[width=1.1\textwidth]{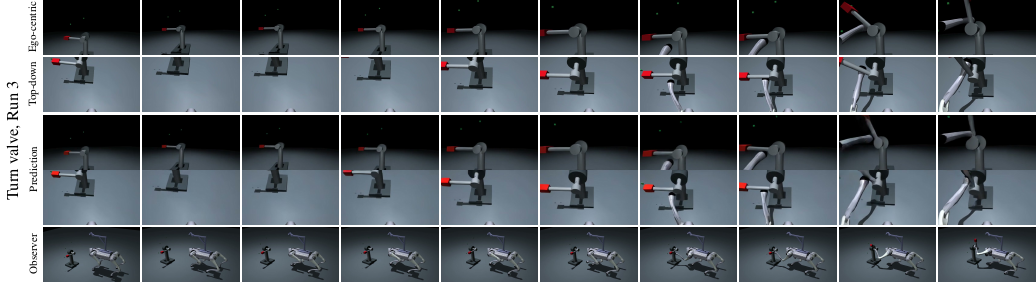}}\\[8pt]
  \makebox[\textwidth][c]{\includegraphics[width=1.1\textwidth]{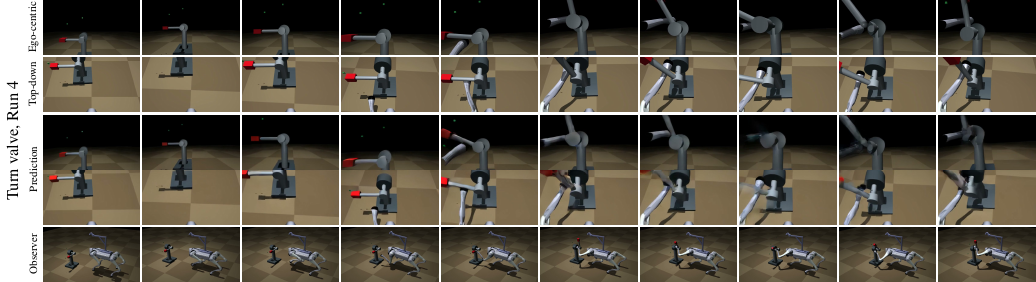}}
  \caption{\textbf{Turning the valve in QLM-Bench.} Runs 1 to 4 from top to bottom.}
  \label{fig:suppl_rollout_valve}
\end{minipage}

\clearpage
\noindent\begin{minipage}{\textwidth}
  \captionsetup{type=figure}
  \centering
  \makebox[\textwidth][c]{\includegraphics[width=1.1\textwidth]{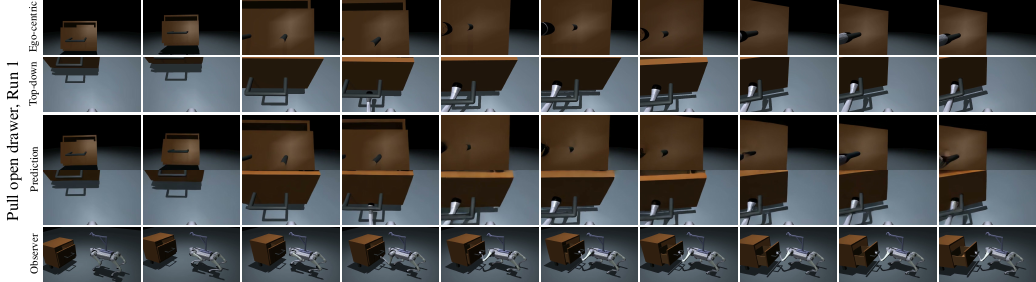}}\\[8pt]
  \makebox[\textwidth][c]{\includegraphics[width=1.1\textwidth]{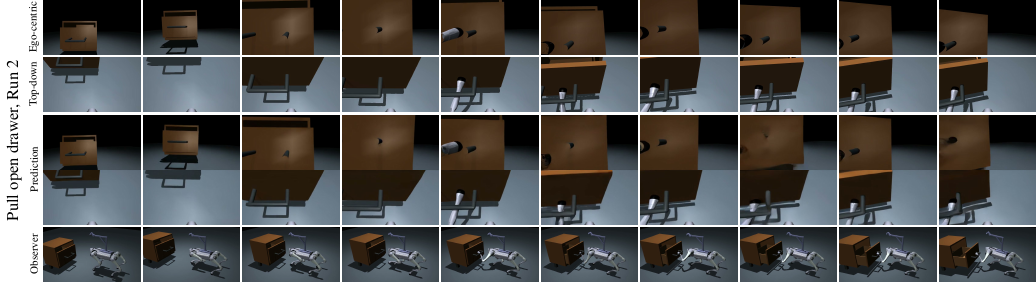}}\\[8pt]
  \makebox[\textwidth][c]{\includegraphics[width=1.1\textwidth]{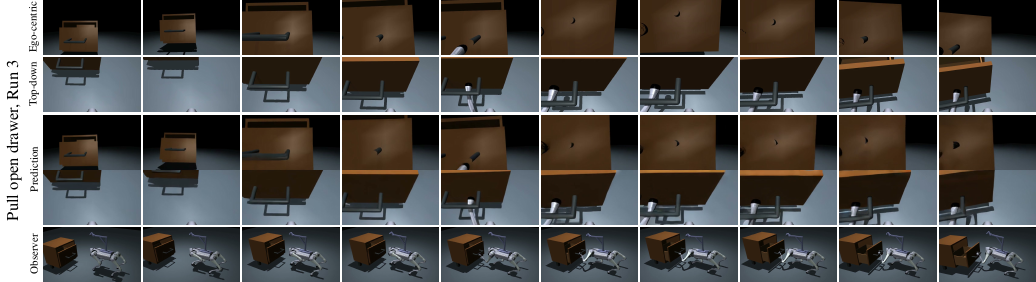}}\\[8pt]
  \makebox[\textwidth][c]{\includegraphics[width=1.1\textwidth]{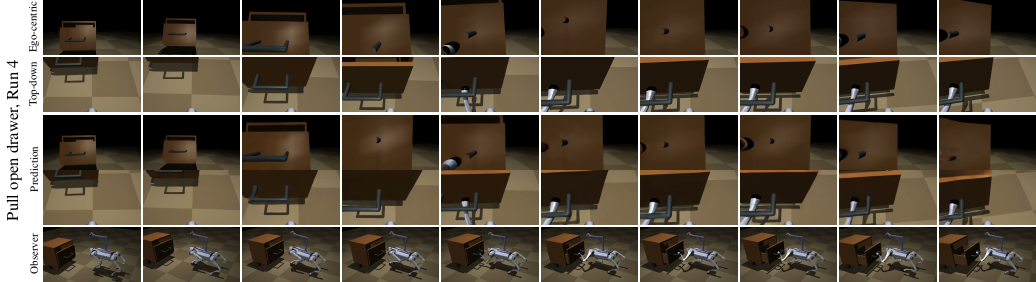}}
  \caption{\textbf{Pulling the drawer open in QLM-Bench.} Runs 1 to 4 from top to bottom.}
  \label{fig:suppl_rollout_drawer}
\end{minipage}

\clearpage
\noindent\begin{minipage}{\textwidth}
  \captionsetup{type=figure}
  \centering
  \makebox[\textwidth][c]{\includegraphics[width=1.1\textwidth]{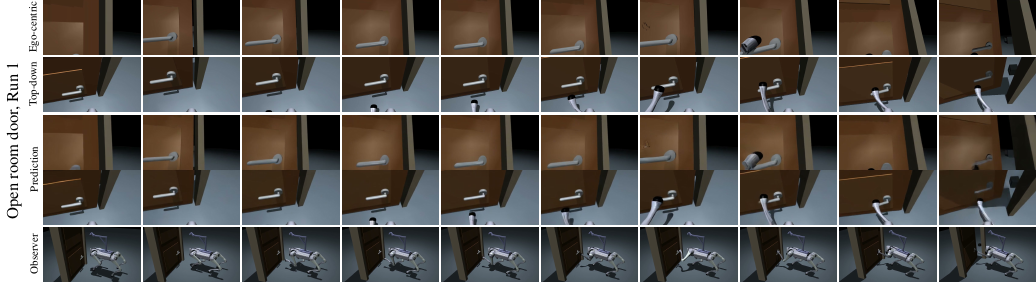}}\\[8pt]
  \makebox[\textwidth][c]{\includegraphics[width=1.1\textwidth]{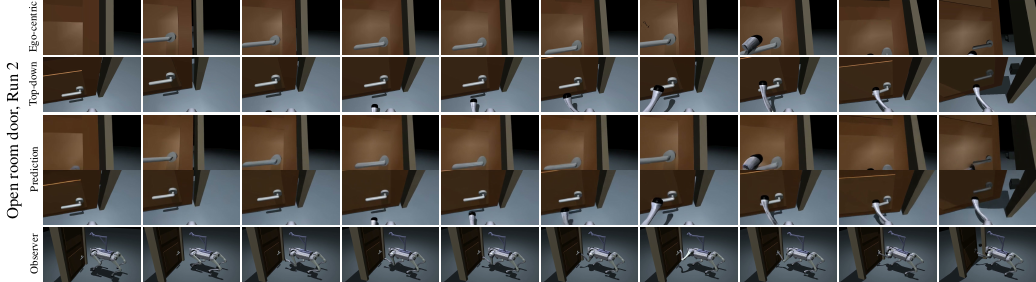}}\\[8pt]
  \makebox[\textwidth][c]{\includegraphics[width=1.1\textwidth]{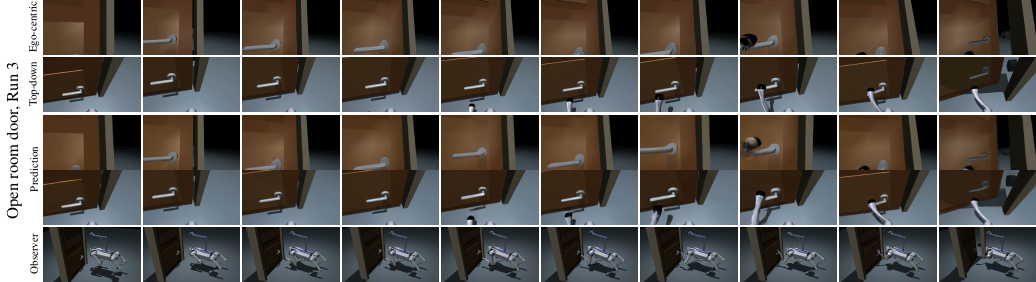}}\\[8pt]
  \makebox[\textwidth][c]{\includegraphics[width=1.1\textwidth]{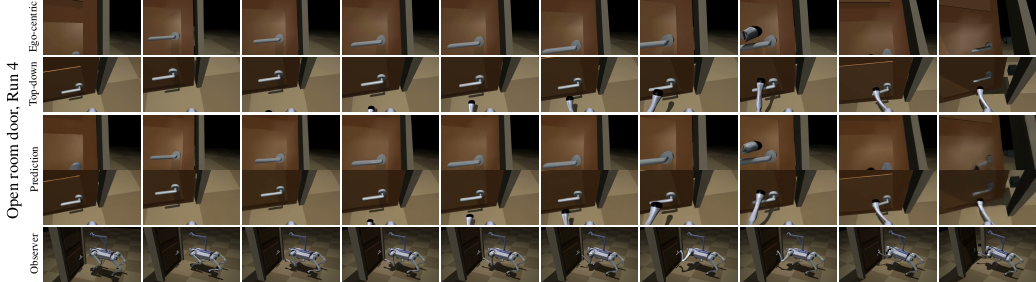}}
  \caption{\textbf{Opening the room door in QLM-Bench.} Runs 1 to 4 from top to bottom.}
  \label{fig:suppl_rollout_door}
\end{minipage}

\clearpage
\noindent\begin{minipage}{\textwidth}
  \captionsetup{type=figure}
  \centering
  \makebox[\textwidth][c]{\includegraphics[width=1.1\textwidth]{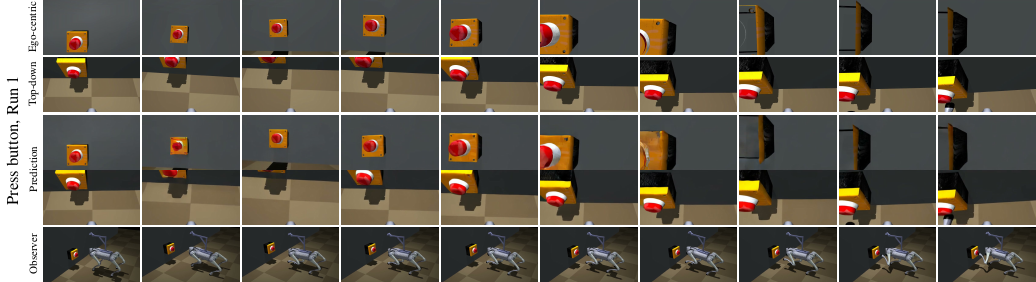}}\\[8pt]
  \makebox[\textwidth][c]{\includegraphics[width=1.1\textwidth]{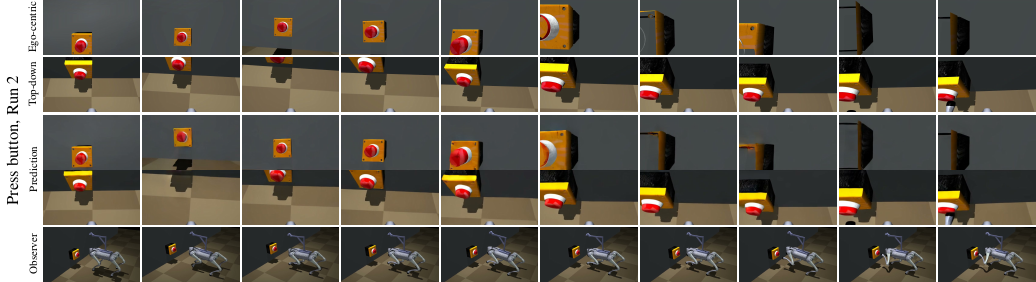}}\\[8pt]
  \makebox[\textwidth][c]{\includegraphics[width=1.1\textwidth]{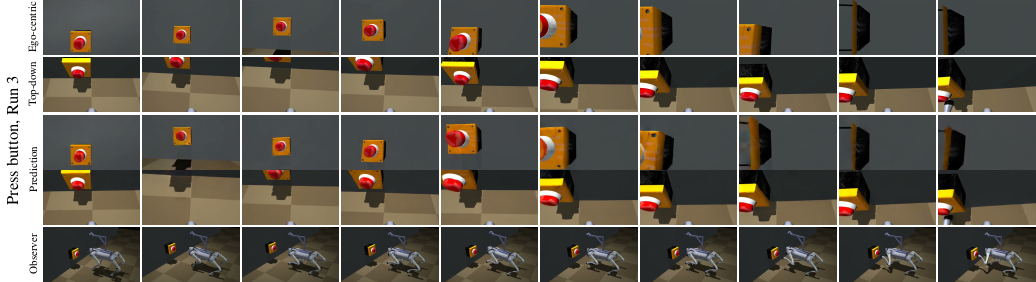}}\\[8pt]
  \makebox[\textwidth][c]{\includegraphics[width=1.1\textwidth]{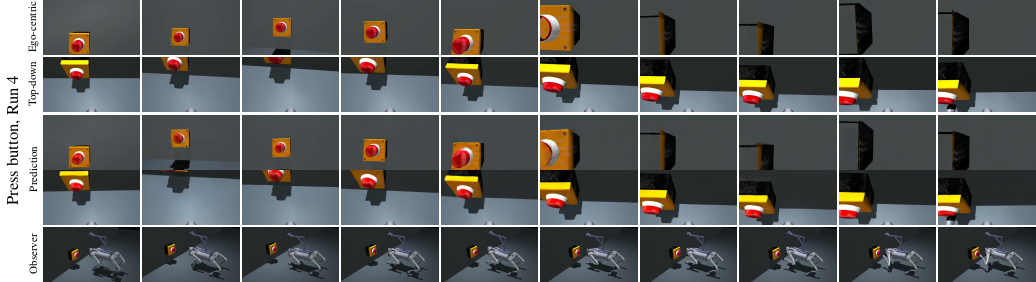}}
  \caption{\textbf{Pressing the button in QLM-Bench.} Runs 1 to 4 from top to bottom.}
  \label{fig:suppl_rollout_press}
\end{minipage}

\clearpage
\noindent\begin{minipage}{\textwidth}
  \captionsetup{type=figure}
  \centering
  \makebox[\textwidth][c]{\includegraphics[width=1.1\textwidth]{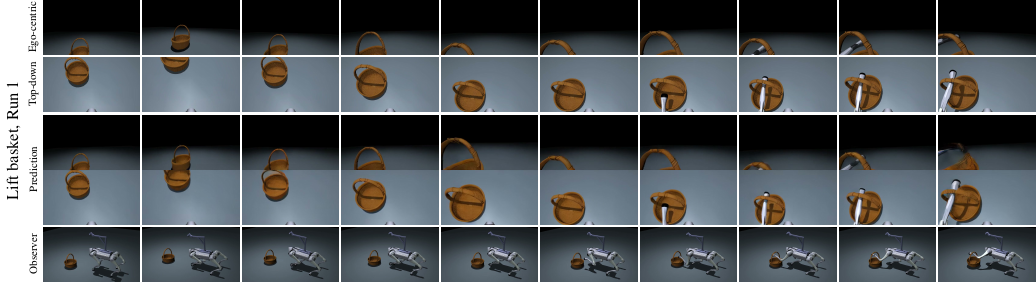}}\\[8pt]
  \makebox[\textwidth][c]{\includegraphics[width=1.1\textwidth]{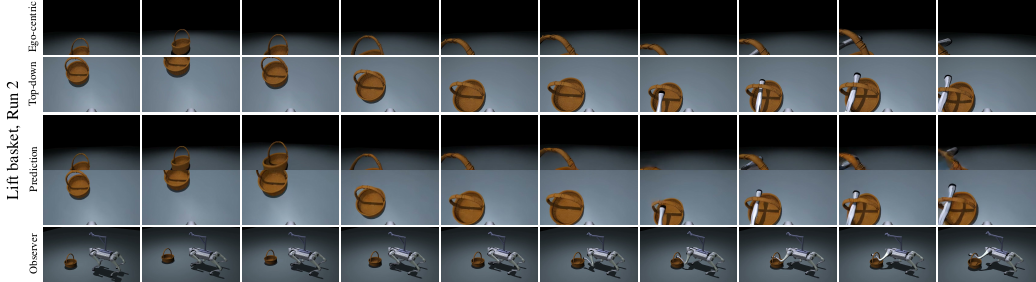}}\\[8pt]
  \makebox[\textwidth][c]{\includegraphics[width=1.1\textwidth]{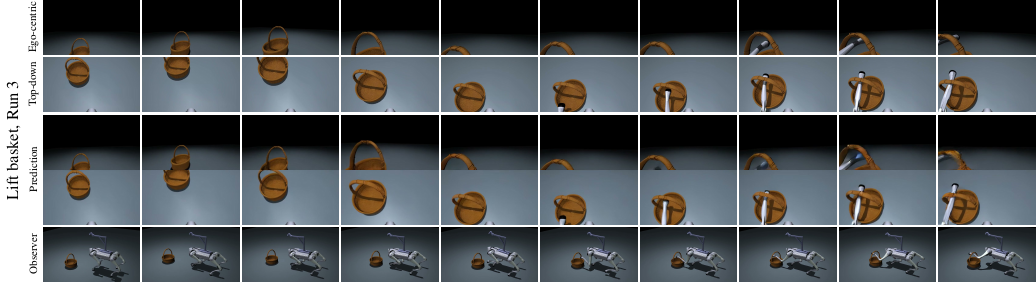}}\\[8pt]
  \makebox[\textwidth][c]{\includegraphics[width=1.1\textwidth]{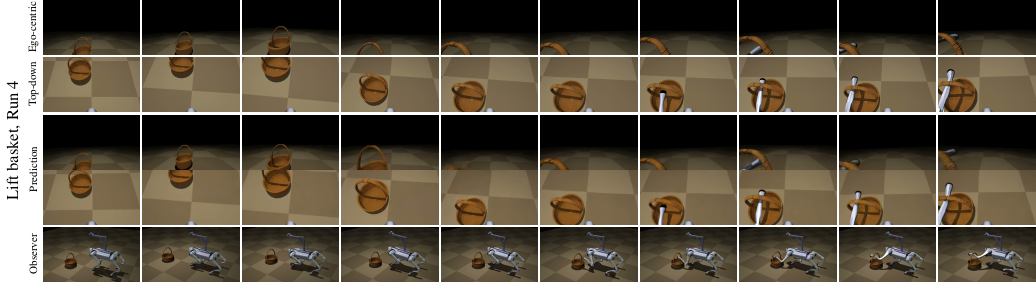}}
  \caption{\textbf{Lifting the basket in QLM-Bench.} Runs 1 to 4 from top to bottom.}
  \label{fig:suppl_rollout_basket}
\end{minipage}

\clearpage
\noindent\begin{minipage}{\textwidth}
  \captionsetup{type=figure}
  \centering
  \makebox[\textwidth][c]{\includegraphics[width=1.1\textwidth]{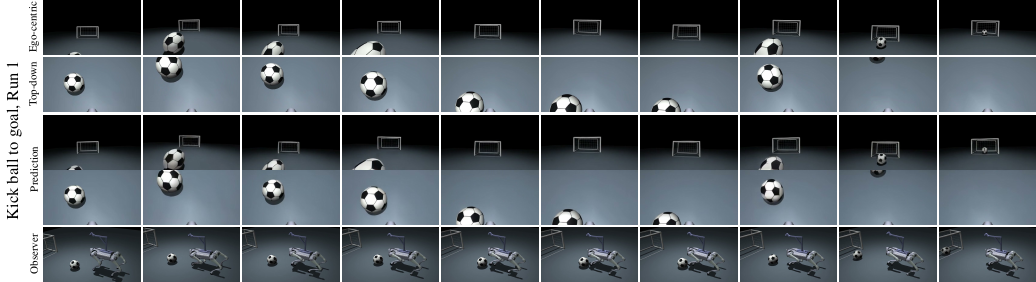}}\\[8pt]
  \makebox[\textwidth][c]{\includegraphics[width=1.1\textwidth]{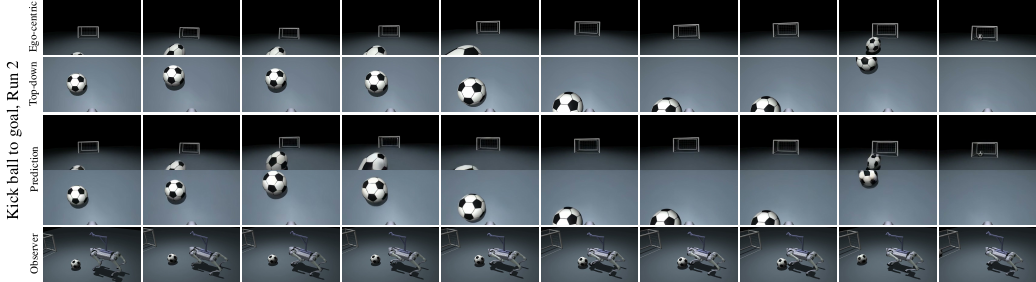}}\\[8pt]
  \makebox[\textwidth][c]{\includegraphics[width=1.1\textwidth]{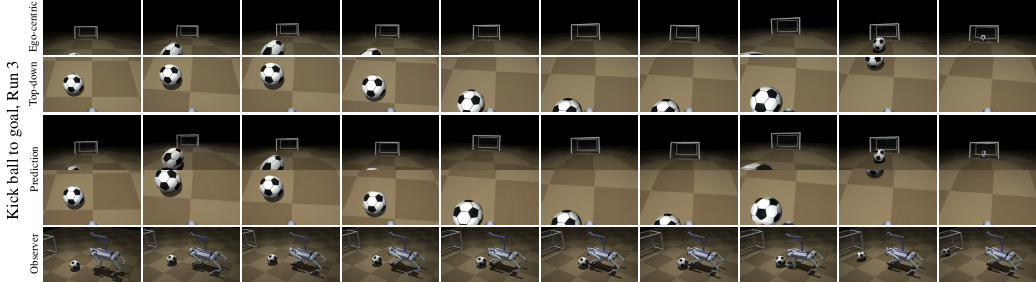}}\\[8pt]
  \makebox[\textwidth][c]{\includegraphics[width=1.1\textwidth]{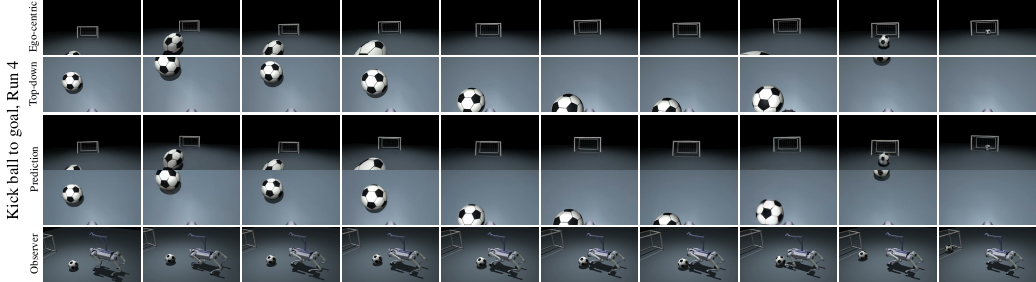}}
  \caption{\textbf{Kicking the ball through the goal in QLM-Bench.} Runs 1 to 4 from top to bottom.}
  \label{fig:suppl_rollout_kick}
\end{minipage}

\clearpage
\noindent\begin{minipage}{\textwidth}
  \captionsetup{type=figure}
  \centering
  \makebox[\textwidth][c]{\includegraphics[width=1.1\textwidth]{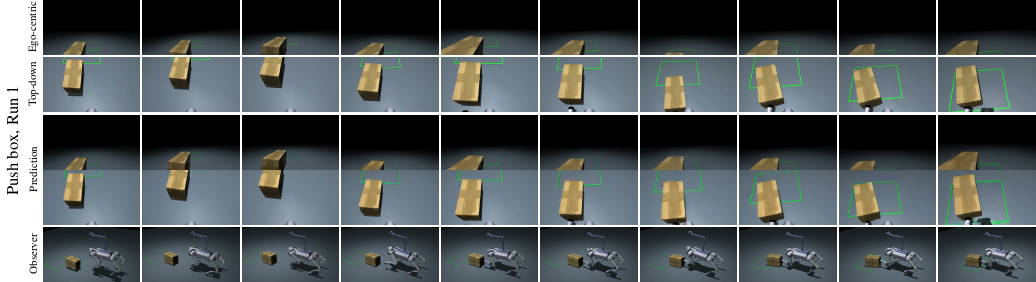}}\\[8pt]
  \makebox[\textwidth][c]{\includegraphics[width=1.1\textwidth]{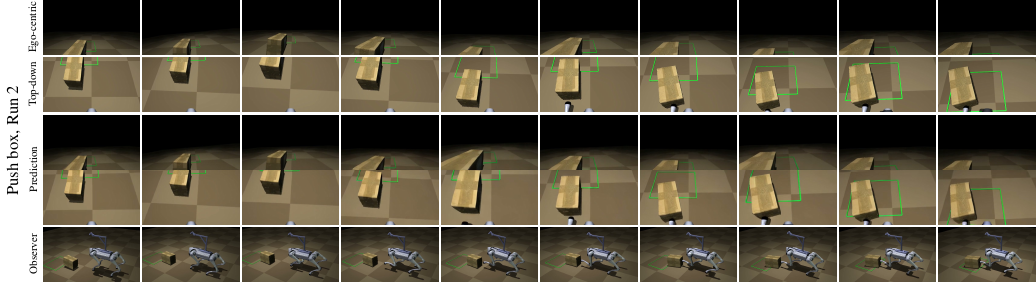}}\\[8pt]
  \makebox[\textwidth][c]{\includegraphics[width=1.1\textwidth]{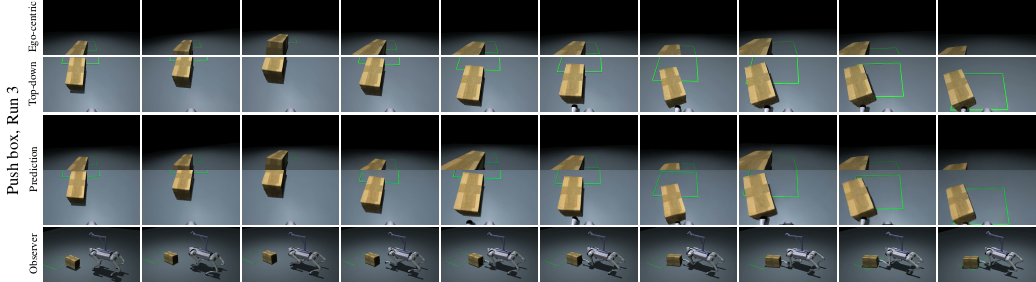}}\\[8pt]
  \makebox[\textwidth][c]{\includegraphics[width=1.1\textwidth]{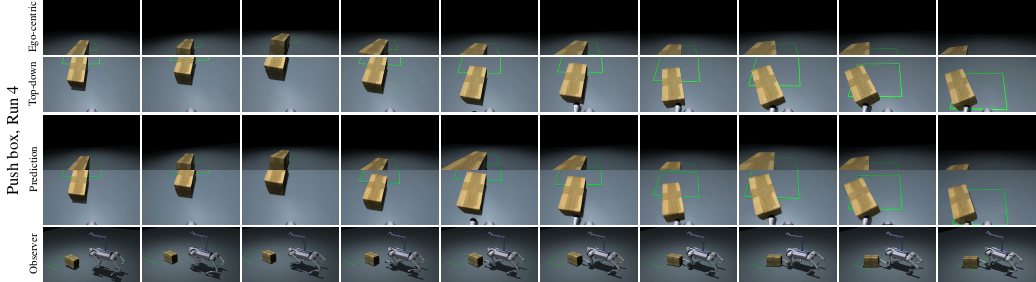}}
  \caption{\textbf{Pushing the box into the target area in QLM-Bench.} Runs 1 to 4 from top to bottom.}
  \label{fig:suppl_rollout_push}
\end{minipage}

\clearpage
\noindent\begin{minipage}{\textwidth}
  \captionsetup{type=figure}
  \centering
  \makebox[\textwidth][c]{\includegraphics[width=1.1\textwidth]{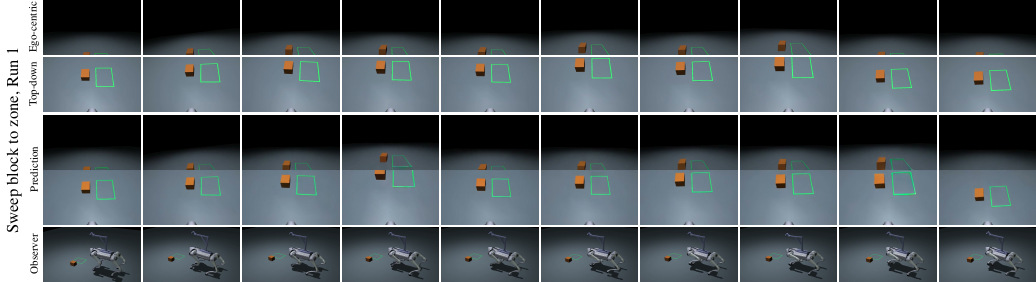}}\\[8pt]
  \makebox[\textwidth][c]{\includegraphics[width=1.1\textwidth]{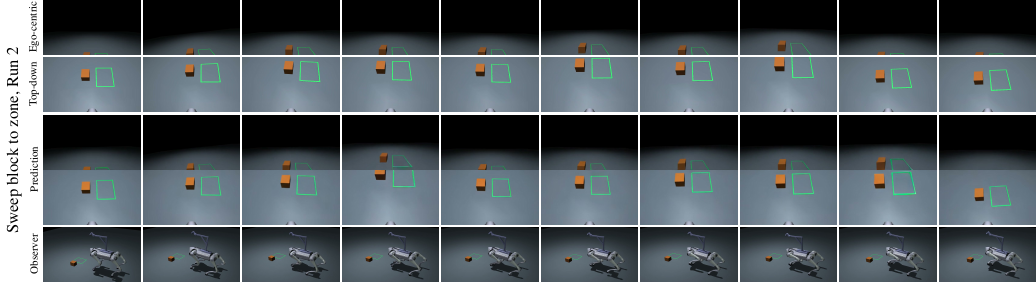}}\\[8pt]
  \makebox[\textwidth][c]{\includegraphics[width=1.1\textwidth]{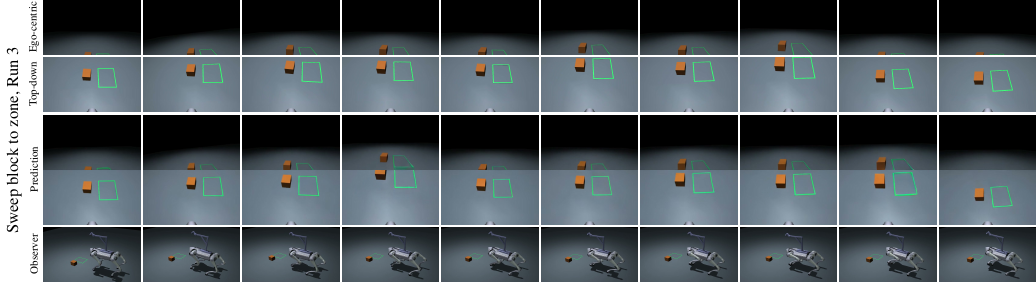}}\\[8pt]
  \makebox[\textwidth][c]{\includegraphics[width=1.1\textwidth]{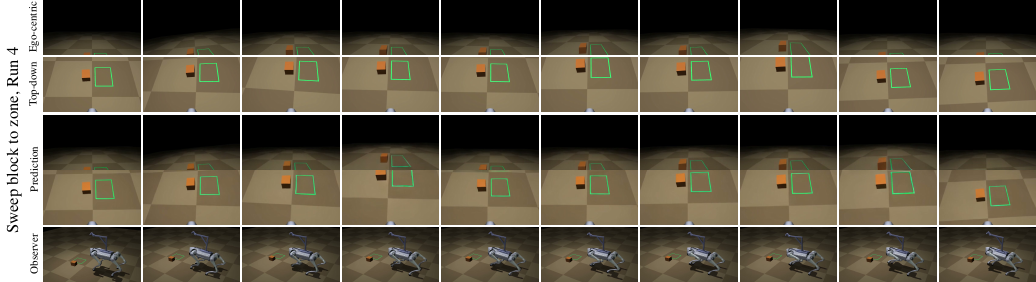}}
  \caption{\textbf{Sweeping the block into the target area in QLM-Bench.} Runs 1 to 4 from top to bottom.}
  \label{fig:suppl_rollout_sweep}
\end{minipage}

\end{document}